\documentclass[10pt,journal,compsoc]{IEEEtran}

\makeatletter
\DeclareRobustCommand\onedot{\futurelet\@let@token\@onedot}
\def\@onedot{\ifx\@let@token.\else.\null\fi\xspace}

\def\eg{\emph{e.g}\onedot}

\makeatother
\usepackage{xspace}

\ifCLASSOPTIONcompsoc
  \usepackage[nocompress]{cite}
\else
  \usepackage{cite}
\fi
\ifCLASSINFOpdf
\else
\fi

\usepackage{hyperref}       
\usepackage{url}            
\usepackage{booktabs}       
\usepackage{nicefrac}       
\usepackage{microtype}      
\usepackage{subfigure}
\usepackage{makecell}

\usepackage{enumitem}
\usepackage{amsmath}
\usepackage{enumerate}
\usepackage{xcolor}
\usepackage{bm}
\usepackage{graphicx}
\usepackage{color}
\usepackage{chngpage}
\usepackage{import}
\usepackage{amssymb,amsfonts}
\usepackage{textcomp}
\usepackage{flushend}
\usepackage{hhline}
\usepackage{multirow}
\usepackage{array}
\usepackage{gensymb}
\usepackage{nameref}
\usepackage{tabularx}
\usepackage[normalem]{ulem}
\usepackage{algorithm}
\usepackage{algpseudocode}
\usepackage[figuresight]{rotating}
\usepackage{makecell}

\makeatletter
\algrenewcommand\ALG@beginalgorithmic{\small}
\algrenewcommand\algorithmiccomment[2][\small]{{#1\hfill\(\triangleright\) #2}}
\makeatother
\usepackage{enumitem}

\DeclareRobustCommand\nobreakspace{\leavevmode\nobreak\ }
\newcolumntype{"}{!{\vrule width 1pt}}
\newcolumntype{L}[1]{>{\raggedright\let\newline\\\arraybackslash\hspace{0pt}}m{#1}}
\newcolumntype{C}[1]{>{\centering\let\newline\\\arraybackslash\hspace{0pt}}m{#1}}
\newcolumntype{R}[1]{>{\raggedleft\let\newline\\\arraybackslash\hspace{0pt}}m{#1}}

\hypersetup{
    colorlinks=true,
    linkcolor=red,
    citecolor=cyan,
    filecolor=magenta,      
    urlcolor=cyan,
    }

\begin{document}
\title{Is Sora a World Simulator? A Comprehensive Survey on General World Models and Beyond}

\author{\normalsize{
Zheng Zhu$^*$, Xiaofeng Wang$^*$, Wangbo Zhao$^*$, Chen Min$^*$, Bohan Li$^*$, Nianchen Deng$^*$, Min Dou$^*$, \\
 Yuqi Wang$^*$, Botian Shi$^\dag$, Kai Wang$^\dag$, Chi Zhang$^\dag$, Yang You$^\dag$, Zhaoxiang Zhang$^\dag$, \\
 Dawei Zhao$^\dag$, Liang Xiao$^\dag$,
 Jian Zhao$^\dag$,
Jiwen Lu$^\dag$,  Guan Huang$^\dag$

\IEEEcompsocitemizethanks{
\IEEEcompsocthanksitem $^*$ indicates equal contributions. $^\dag$ indicates corresponding authors.
\IEEEcompsocthanksitem Zheng Zhu and Guan Huang are with GigaAI, Beijing, China.
\IEEEcompsocthanksitem Xiaofeng Wang, Yuqi Wang and Zhaoxiang Zhang are with Institute of Automation, Chinese Academy of Sciences, Beijing, China.
\IEEEcompsocthanksitem Wangbo Zhao, Kai Wang and Yang You are with National University of Singapore, Singapore.
\IEEEcompsocthanksitem Bohan Li is with Shanghai Jiao Tong University, Shanghai, China.
\IEEEcompsocthanksitem Chen Min is with Institute of Computing Technology, Chinese Academy of Sciences, Beijing, China.
\IEEEcompsocthanksitem Nianchen Deng, Min Dou and Botian Shi are with Shanghai Artificial Intelligence Laboratory, Shanghai, China.
\IEEEcompsocthanksitem Chi Zhang is with Mach Drive, Beijing, China.
\IEEEcompsocthanksitem Dawei Zhao and Liang Xiao are with Defense Innovation Institute, Beijing, China.
\IEEEcompsocthanksitem Jian Zhao is with EVOL Lab, Institute of AI, China Telecom, and Northwestern Polytechnical University.
\IEEEcompsocthanksitem Jiwen Lu is with Tsinghua University, Beijing, China.

}
}
}

\markboth{IEEE TRANSACTIONS ON Pattern Analysis and Machine Intelligence, VOL. XX, NO. XX, XXX 2025}%
{Shell \MakeLowercase{\textit{et al.}}: }

\IEEEtitleabstractindextext{%
\begin{abstract}
General world models represent a crucial pathway toward achieving Artificial General Intelligence (AGI), serving as the cornerstone for various applications ranging from virtual environments to decision-making systems. Recently, the emergence of the Sora model has attained significant attention due to its remarkable simulation capabilities, which exhibits an incipient comprehension of physical laws. In this survey, we embark on a comprehensive exploration of the latest advancements in world models. Our analysis navigates through the forefront of generative methodologies in video generation, where world models stand as pivotal constructs facilitating the synthesis of highly realistic visual content. Additionally, we scrutinize the burgeoning field of autonomous-driving world models, meticulously delineating their indispensable role in reshaping transportation and urban mobility. Furthermore, we delve into the intricacies inherent in world models deployed within autonomous agents, shedding light on their profound significance in enabling intelligent interactions within dynamic environmental contexts. At last, we examine challenges and limitations of world models, and discuss their potential future directions. We hope this survey can serve as a foundational reference for the research community and inspire continued innovation.
This survey will be regularly updated at: \url{https://github.com/GigaAI-research/General-World-Models-Survey}.
\end{abstract}

\begin{IEEEkeywords}
World models, Generative models, Video generation, Autonomous driving, Autonomous agents
\end{IEEEkeywords}}

\maketitle

\IEEEdisplaynontitleabstractindextext

\IEEEpeerreviewmaketitle

\section{Introduction}
\IEEEPARstart{I}n the pursuit of Artificial General Intelligence (AGI), the development of general world models stands as a fundamental avenue. General world models seek to understand the world through generative processes. Notably, the introduction of the Sora model \cite{videoworldsimulators2024} has garnered significant attention. Its remarkable simulation capabilities not only demonstrate an initial comprehension of physical laws but also highlight the promising advancements in world models. As we stand at the forefront of AI-driven innovation, it is crucial to delve deeper into the realm of world models, unraveling their complexities, evaluating their current developmental stage, and contemplating the potential trajectories they may follow in the future.

\begin{figure*}[t]
	\centering
	\includegraphics[width=0.8\textwidth]{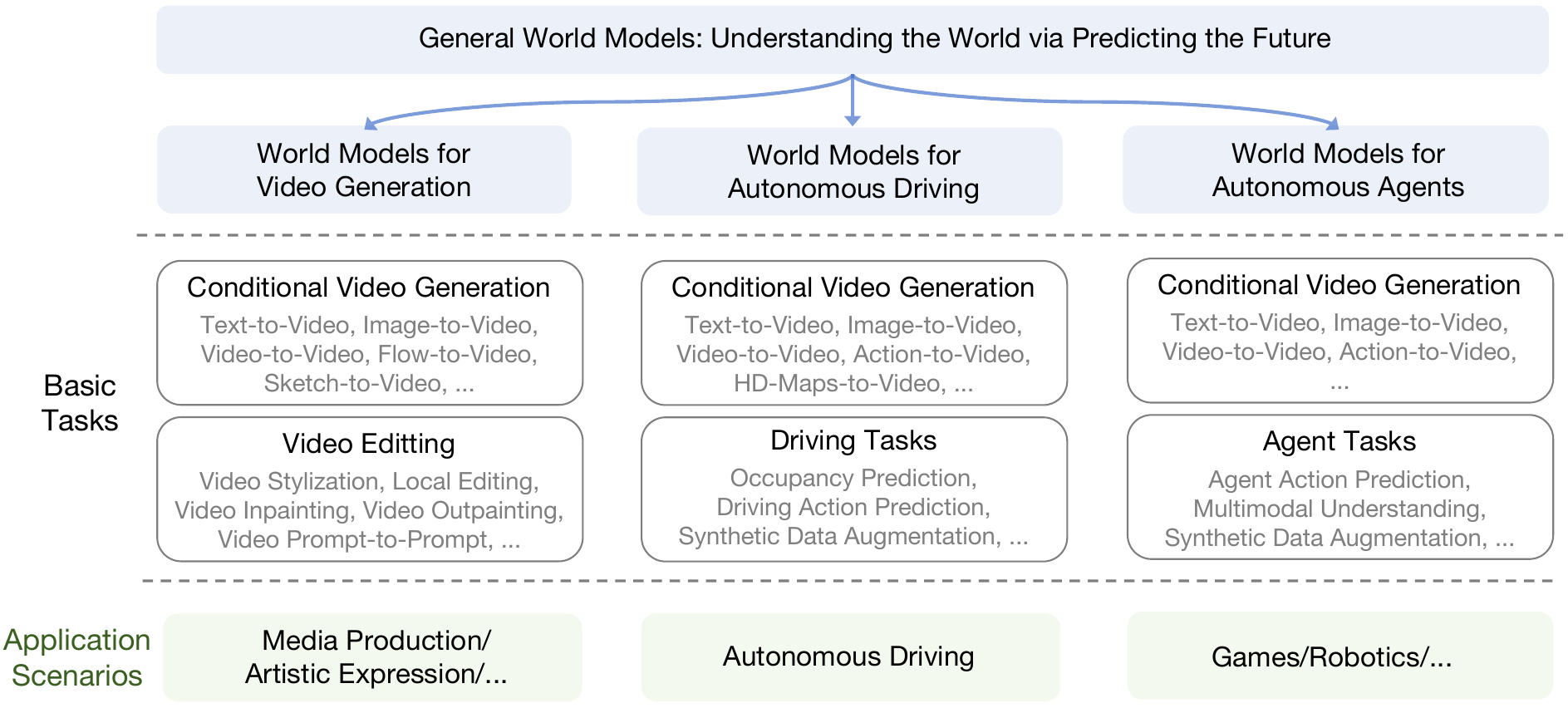} 
	\caption{This survey focuses on world models for video generation, world models for autonomous driving, and world models for autonomous agents. Video generation world models specialize in conditional video generation and various video editing tasks. These video generation techniques aid in the understanding of complex scenes and decision-making processes in autonomous driving and autonomous agents world models. The applications of these world models are broad, ranging from media production and artistic expression to action prediction in autonomous driving and agent systems.}
	\label{fig:intro}
\end{figure*}

World models predict the future to grow comprehension of the world. This predictive capacity holds immense promise for video generation, autonomous driving, and the development of autonomous agents, which represent three mainstream directions of development in world models.
As shown in Figure~\ref{fig:intro}, video generation world models encompass the generation and editing of videos to understand and simulate the world, which are valuable for media production and artistic expression. 
Autonomous driving world models, aided by techniques of video generation, create driving scenarios and learn driving elements and policies from driving videos. This knowledge assists in generating driving actions directly or training driving policy networks, aiding in end-to-end autonomous driving.
Similarly, agent world models utilize video generation to establish intelligent interactions in dynamic environments. Unlike driving models, they build policy networks applicable to various contexts, either virtual (e.g., programs in games or simulated environments) or physical (e.g., robots).

Building upon the foundation of comprehensive world modeling, video generation methods unveil physical laws through visual synthesis.
Initially, the focus of generative models was primarily on image generation \cite{rombach2022high,podell2023sdxl,vqdiff,cogview,ramesh2022hierarchical,glide,unidiff,parti,chen2020generative} and editing \cite{controlnet,gligen,t2i,composer}, laying the foundation for more sophisticated advancements in synthesizing dynamic visual sequences. Over time, generative models \cite{yan2021videogpt,emuvideo,pixeldance,svd,ho2022imagen,gupta2023photorealistic,ho2022imagen,videoldm,kondratyuk2023videopoet,esser2023structure} have evolved to not only capture the static attributes of images, but also seamlessly string together sequences of frames. These models have developed some understanding of physics and motion, which represent early and limited forms of general world models \cite{generalworldmodels}. Notably, 
at the forefront of this evolution stands the Sora model \cite{videoworldsimulators2024}.
By harnessing the power of generative techniques, Sora demonstrates a profound ability to generate intricate visual narratives that adhere to the fundamental principles of the physical world. 
The relationship between generative models and world modeling is symbiotic, with each informing and enriching the other. Generative models can construct vast amounts of data in a controlled environment, which alleviates the need for extensive real-world data collection, particularly beneficial for training AI systems essential in real-world applications.
Moreover, the efficacy of generative models critically hinges upon the depth of comprehension provided by world models. It is the comprehensive understanding of underlying environmental dynamics afforded by world models that empowers generative models to produce visually compelling signals of superior quality while adhering to stringent physical constraints. Thereby enhancing their realism and utility in various domains.

The ability of world models to understand the environment not only enhances video generation quality, but also benefits real-world driving scenarios. By employing predictive techniques to comprehend driving environments, world models are reshaping transportation and urban mobility by anticipating future driving scenarios, thereby enhancing safety and efficiency.
World methods, aimed at establishing dynamic models of environments, are crucial in autonomous driving, where precise predictions about the future are essential for safe maneuvering. However, constructing world models for autonomous driving presents unique challenges, primarily due to the sample complexity inherent in real-world driving scenarios. Early methods \cite{mile,isodream,sem2} attempt to address these challenges by reducing the search space and incorporating explicit disentanglement of visual dynamics. Despite progress, a critical limitation lies in the predominant focus on simulation environments.
Recent advances have seen autonomous driving world models leverage generative models to tackle real-world scenarios with larger search spaces. GAIA-1 \cite{gaia} employs a Transformer to predict the next visual token, effectively constructing the driving world model. This approach enables anticipating multiple potential futures based on various prompts, such as weather conditions, scenes, traffic participants, and vehicle actions. Similarly, methods like DriveDreamer \cite{drivedreamer} and Panacea \cite{wen2023panacea} leverage pre-trained diffusion models to learn driving world models from real-world driving videos. These techniques harness the structured information inherent in driving scenes to controllably generate high-quality driving videos, which can even enhance training for driving perception tasks. DriveDreamer2 \cite{drivedreamer2}, based on DriveDreamer, further integrates large language models to enhance the performance of driving world models and user interaction. It enables the generation of controllable driving scene videos solely through natural language input, encompassing even rare scenarios like sudden overtaking maneuvers. Furthermore, Drive-WM \cite{drivewm} demonstrates the feasibility of directly training end-to-end driving using generated driving scene videos, significantly improving end-to-end driving performance. By anticipating future scenarios, these models empower vehicles to make informed decisions, ultimately leading to safer and more efficient navigation on the roads. Moreover, this integration not only improves transportation systems' safety and efficiency but also opens new possibilities for urban planning and design.

Beyond their established utility in driving scenarios, world models have increasingly become integral to the functioning of autonomous agents, facilitating intelligent interactions across a myriad of contexts. For instance, world models in game agents not only augment the gaming experience but also propel the development of sophisticated game algorithms. The Dreamer series \cite{hafner2020dreamerV1,hafner2021dreamerV2,hafner2023dreamerv3} exemplify this with its adept use of world models to predict future states within gaming environments. This capability enables game agents to learn in imagination, markedly decreasing the necessary volume of interactions for effective learning. In robotic systems, innovative approaches further underscore the versatility and potential of world models. UniPi~\cite{du2024learning}, for instance, reimagines the decision-making problem in robotics as a text-to-video task. Its policy-as-video formulation fosters learning and generalization across diverse robot manipulation tasks. Similarly, UniSim \cite{yang2024unisim} introduces a simulator of dynamic interactions through generative modeling, which can then be deployed in real-world scenarios without prior exposure. RoboDreamer \cite{zhou2024robodreamer} pushes the envelope by leveraging world models to propose plans involving combinations of actions and objects, thus solving unprecedented tasks in novel robotic execution environments. The multifaceted applications of world models extend beyond games and robotics. LeCun's proposal of the Joint-Embedding Predictive Architecture (JEPA) \cite{lecun2022jepa} heralds a significant departure from traditional generative models. JEPA learns to map input data to predicted outputs within a higher-level representation space, which enables the model to concentrate on learning more semantic features, enriching its capability for understanding and predicting across various modalities. 
  

Based on the comprehensive discussions presented above, it is evident that research on world models holds tremendous potential towards achieving AGI and has wide-ranging applications across various domains. Therefore, world models warrant significant attention from both academia and industry, requiring sustained efforts over an extended period.
In comparison to recent surveys \cite{guan2024world,cho2024sora,liu2024sora,sunsora} on world models, our survey offers broader coverage. It not only encompasses generative world models in video generation but also delves into the applications of world models in decision-making systems such as autonomous driving and robotics.
We envision this survey to offer valuable insights for newcomers embarking on their journey into this field, while also stimulating critical thinking and discussion among established researchers in the community.

The main contributions of this survey can be summarized as follows:
(1) We present a holistic examination of recent advancements in world model research, encompassing profound philosophical perspectives and detailed discussions.
(2) Our analysis delves deeply into the literature surrounding world models for video generation, autonomous driving, and autonomous agents, uncovering their applications in media production, artistic expression, end-to-end driving, games, and robots.
(3) We assess the existing challenges and limitations of world models and delve into prospective avenues for future research, with the intention of steering and igniting further progress in world models.

\section{Video Generation as a General World Model}
The video generation task aims to create various realistic videos, requiring the model to understand and simulate the mechanism in the physical world, which aligns with the objective of building a general world model. In this section, we first introduce the technologies behind the video generation models in Section~\ref{sec:tech_video}. Then, in Section~\ref{sec:adv_video}, we present and review the advanced video generation models emerging in recent years. Finally, we discuss Sora in Section~\ref{sec:sora}, which is considered to be the largest breakthrough in video generation. 

\subsection{Technologies behind Video Generation} 
\label{sec:tech_video}
The concept of video generation contains several different tasks based on the conditions, such as class, text, or image. This survey mainly focuses on the scenario where the text condition is given, known as text-to-video generation. In this section, we first briefly introduce the visual foundation models, which are widely used in generation models. Then, we
present the text encoders for extracting text features from the text condition. Finally, we review the evolution of generation techniques. 

\subsubsection{Visual Foundation Models} \label{sec:visual_found}
The visual foundation models were originally proposed to tackle traditional computer vision tasks, for example, image classification \cite{deng2009imagenet}, whereas they also inspire the development of generation models. Based on the architecture, they can be roughly categorized into convolution-based models and Transformer-based models, both of which can also be extended to the video data. 

\noindent
\textbf{Convolution-based Models.}
The convolution-based models for vision tasks have been fully explored in the last decades. Staring from LeNet \cite{lecun1998gradient}, AlexNet \cite{krizhevsky2012imagenet}, VGGNet \cite{simonyan2014very}, InceptionNet \cite{szegedy2015going}, ResNet \cite{he2016deep}, DenseNet \cite{huang2017densely} are gradually proposed to tackle the image recogntion problems. These models are adopted as a backbone model for other visual tasks \cite{ren2015faster, he2017mask, ronneberger2015u}. Typically, U-Net \cite{ronneberger2015u} builds a U-shape architecture based on a backbone model for image segmentation tasks. The U-shape architecture enables the model can leverage both the low-level and high-level features from the backbone, which significantly improves the pixel-wise prediction. Benefiting from the superiority of pixel-wise prediction,  the U-shape architecture is also widely used in image generation models \cite{ho2020denoising, dhariwal2021diffusion, rombach2022high}. 

\noindent
\textbf{Transformer-based Models.}
The Transformer is proposed in \cite{vaswani2017attention} for machine translation tasks and applied to vision recognition by ViT \cite{dosovitskiy2020image}. In ViT, images are divided into patches, then projected into tokens and finally processed by a series of multi-head self-attention and multi-layer perceptron blocks. Its ability to capture long-range dependencies in images enables its superiority in image recognition. After that, distillation \cite{touvron2021training}, window-attention \cite{liu2021swin}, and mask image modeling \cite{bao2021beit, he2022masked} approaches are introduced to improve the training or inference efficiency of vision Transformers. Except for the success in image recognition, Transformer-based models also demonstrate superiority in various visual tasks, such as object detection \cite{carion2020end, zhu2020deformable, yao2021efficient, zhang2022dino}, semantic segmentation \cite{xie2021segformer, strudel2021segmenter, zheng2021rethinking}, and image generation \cite{peebles2023scalable, bao2023all, hatamizadeh2023diffit}. Thanks to its good scalability property, the Transformer-based model DiT \cite{peebles2023scalable} has become the main architecture of Sora.  

\noindent
\textbf{Extension to Video.}
The methods mentioned above are mainly designed for image data. Researchers further extend these methods to solve problems in the video domain. Convolution-based models \cite{ji20123d, tran2015learning, carreira2017quo, tran2017convnet, feichtenhofer2019slowfast, feichtenhofer2020x3d, yang2020temporal} usually introduce 3D convolution layers to building the spatial-temporal relationships in video data. Transformer-based methods \cite{arnab2021vivit, bertasius2021space, li2022uniformer, liu2022video} extend and improve the multi-head self-attention from spatial-only design to jointly modeling spatial-temporal relationships. These methods also inspire the architecture design of text-to-video generation models, such as \cite{yu2023magvit, yu2023language, kondratyuk2023videopoet}. 

\subsubsection{Text Encoders}
The text encoder is adopted to extract the text embedding for a given text prompt in image or video generation. Existing generation methods usually employ the text encoder of a multi-modal model or directly use a language model to conduct the embedding extraction. In the following, we will briefly present representative multi-modal models and language models. 

\noindent 
\textbf{Pre-trained Multi-modal Models.}
The pre-trained multi-modal models, such as \cite{li2021align, radford2021learning, li2022blip}, align the representation of image and text in the embedding space. It usually consists of an image encoder as well as a text encoder, which naturally can be adapted to inject text information into generation models. CLIP  \cite{radford2021learning}  is a typical pre-trained multi-modal model, which has been widely used in image/video generation models \cite{podell2023sdxl, ramesh2022hierarchical, rombach2022high, svd}. It is pre-trained with large-scale image-text pairs through contrastive learning \cite{jaiswal2020survey} and demonstrates superior performance across various tasks.  However, CLIP is pre-trained for image-text alignment instead of comprehending complex text prompts. This drawback may limit the generation performance when the given prompt is long and detailed. 

\noindent
\textbf{Pre-trained Language Models.}
The pre-trained language models are usually pre-trained on the large-scale corpus, thus having transferable ability on various downstream language tasks. BERT \cite{devlin2018bert} is an early attempt at language model pre-training, which designed several tasks to push the model learning from unlabeled data. This paradigm also inspires follow-up works, such as RoBERTa \cite{liu2019roberta} and BART \cite{lewis2019bart}. With the increasing model size and enlarging training dataset, the pre-trained models demonstrate surprising abilities, which are usually named as larger language models (LLMs) \cite{radford2018improving, radford2019language, 
brown2020language, raffel2020exploring, touvron2023llama, touvron2023llama2, achiam2023gpt}. T5 \cite{raffel2020exploring} and Llama-2 \cite{touvron2023llama} are two widely used LLMs in generation tasks \cite{hu2024ella, chen2023pixart, saharia2022photorealistic, wu2023paragraph} since their superior performance and open avaliablity. The LLMs provide a better understanding of long text prompts than CLIP, thus helping the generation to follow the instructions of humans. 

\subsubsection{Generation Techniques} \label{sec:geneartion_tech}
In this section, we review the development of generation techniques in recent decades.



\noindent
\textbf{GAN.}
Before the success of diffusion-based methods, GAN introduced in
\cite{goodfellow2014generative} have always been the mainstream methods in image generation.  It has a generator $G$ and a discriminator $D$. The generator $G$ is adapted to generate an output $G(\mathbf{z})$ from a noised $\mathbf{z}$ sampled from a Gaussian distribution and the discriminator $D$ is employed to classifier the output is real or fake.

From the original definition of GAN \cite{goodfellow2014generative}, the generator $G$ and the discriminator $D$ are trained in an adversarial manner. Specifically, we first train the discriminator $D$.  We input real data $\mathbf{x}$ sampled from a data distribution $p_{\text{data}}(\mathbf{x})$ and generated output $G(\mathbf{z})$ into the discriminator $D$  and it learns to improve the discrimination ability on real and fake samples. This can be formulated as:
\begin{equation}
\ell_D=\mathbb{E}_{\mathbf{x} \sim p_{\text {data }}(\mathbf{x})}[\log D(\mathbf{x})]+\mathbb{E}_{\mathbf{z} \sim p_z(\mathbf{z})}[\log (1-D(G(\mathbf{z})))].
\end{equation}
The discriminator $D$ should maximize the loss $\ell_D$. During this process, the parameters in $G$ are frozen. Then, we train the generator $G$ following:
\begin{equation}
\ell_G=\mathbb{E}_{\mathbf{z} \sim p_z(\mathbf{z})}[\log (1-D(G(\mathbf{z})))] .
\end{equation}
The generator $G$ is trained to minimize the loss $\ell_G$ so that the generated samples can approach the real data. The parameters in $D$ are also not updated during this process.

Following works apply GAN to various tasks related to image generation, such as style transfer \cite{zhu2017unpaired, karras2019style, patashnik2021styleclip, brock2018large}, image editing \cite{zhu2020domain, wang2022high, perarnau2016invertible}, and image inpainting \cite{liu2021pd, demir2018patch}.


\noindent
\textbf{Diffusion.} \label{sec:diffusion}
Diffusion-based methods have started to dominate image generation since the Denoising Diffusion Probabilistic  Model (DDPM) \cite{ho2020denoising}, which learns a reverse process to generate an image from a Gaussian distribution $\mathcal{N}(0, \boldsymbol{I})$. It has two processes: the diffusion process (also known as a forward process) and the denoising process (also known as the reverse process). During the diffusion process, we gradually add small Gaussian noise to an image in $T$ timesteps. Given a image $\mathbf{x}_0$ from the data distribution,  we can obtain $\mathbf{x}_T$ through the cumulative distribution of all previous diffusion processes:
\begin{equation}
q\left(\mathbf{x}_{1: T} \mid \mathbf{x}_0\right):=\prod_{t=1}^T q\left(\mathbf{x}_t \mid \mathbf{x}_{t-1}\right),
\end{equation}
where
\begin{equation}
q\left(\mathbf{x}_t \mid \mathbf{x}_{t-1}\right):=\mathcal{N}\left(\mathbf{x}_t ; \sqrt{1-\beta_t} \mathbf{x}_{t-1}, \beta_t I\right)
\end{equation}
$T$ and $\left[\beta_1, \beta_2, \ldots, \beta_T\right]$ denote the diffusion steps and the pre-defined noise schedule, respectively.
We can also obtain the output at the $t$ timestep through
\begin{equation}
q\left(\mathbf{x}_t \mid \mathbf{x}_0\right):=\mathcal{N}\left(\mathbf{x}_t ; \sqrt{\bar{\alpha}_t} \mathbf{x}_0,\left(1-\bar{\alpha}_t\right) I\right),
\end{equation}
where $\alpha_t:=1-\beta_t$ and $\bar{\alpha}_t:=\prod_{i=0}^t \alpha_i$. Thus, we have
\begin{equation}
\mathbf{x}_t=\sqrt{\bar{\alpha}_t} \mathbf{x}_0+\sqrt{1-\bar{\alpha}_t} \epsilon, \quad \epsilon \sim \mathcal{N}(0,1).
\end{equation}  

The denoising process is the reverse of the diffusion process, enabling us to obtain images from the Gaussian noise. To achieve this, a denoising model $\epsilon_\theta$ learns to predict the noise $\epsilon_t$ added at the timestep $t$ through a simplified loss function, which can be formulated as:
\begin{align}
\ell^{\text {simple }}_t(\theta)&=\mathbb{E}_{\mathbf{x}_0, t, \epsilon_t}\left\|\epsilon_\theta\left(\mathbf{x}_t, t\right)-\epsilon\right\|_2^2 \\
&=\mathbb{E}_{\mathbf{x}_0, t, \epsilon_t}\left\|\epsilon_\theta\left(\sqrt{\bar{\alpha}_t} \mathbf{x}_0+\sqrt{1-\bar{\alpha}_t} \boldsymbol{\epsilon}, t\right)-\epsilon\right\|_2^2 
\end{align}
Then, we can denoise step-by-step through
\begin{equation}
\mathbf{x}_{t-1}=\frac{1}{\sqrt{\alpha_t}}\left(\mathbf{x}_t-\frac{\beta_t}{\sqrt{1-\bar{\alpha}_t}} \epsilon_\theta\left(\mathbf{x}_t, t\right)\right) + \beta_t z, 
\end{equation}
where $\mathbf{z} \sim \mathcal{N}(0, I)$.
Although the generation quality of DDPM is satisfactory, its slow generation speed hinders its broader application. Following works attempt to solve this problem by reducing the denoising steps \cite{song2020denoising,luo2023latent,salimans2022progressive,zheng2023fast,song2023consistency} or accelerating the denoising model \cite{pan2024t,shang2023post,fang2024structural,ma2023deepcache}.  

\noindent
\textbf{Autoregressive Modeling.}
Autoregressive modeling has been explored in both language generation methods \cite{radford2018improving,radford2019language,brown2020language} and image generation tasks \cite{chen2020generative,yu2021diverse,yu2022scaling,lee2022autoregressive}.  Given a sequence of tokens $\left(\mathbf{x}_1, \mathbf{x}_2, \ldots, \mathbf{x}_K\right)$, 
the probability of the $k$-th toeken $\mathbf{x}_k$ only depends on tokens $(\mathbf{x}_1, \mathbf{x}_2, \mathbf{x}_{k-1})$. An autoregressive model $p_{\theta}$ is trained to maximize the likelihood of the current token, which can be formulated as:
\begin{equation}
\ell = \sum_k^{K} \log p_{\theta}\left(\mathbf{x}_k \mid \mathbf{x}_1, \mathbf{x}_2, \ldots, \mathbf{x}_{k-1} \right)
\end{equation}
\noindent

Recently, LVM \cite{bai2023sequential}  scales up the amount of training data to 420 billion tokens and model size to 3 billion parameters, demonstrating an ability for general visual reasoning as well as generation and directing a potential way towards the world model. 


\begin{figure*}[!t]
\centering

  \resizebox{1\linewidth}{!} {
    \includegraphics{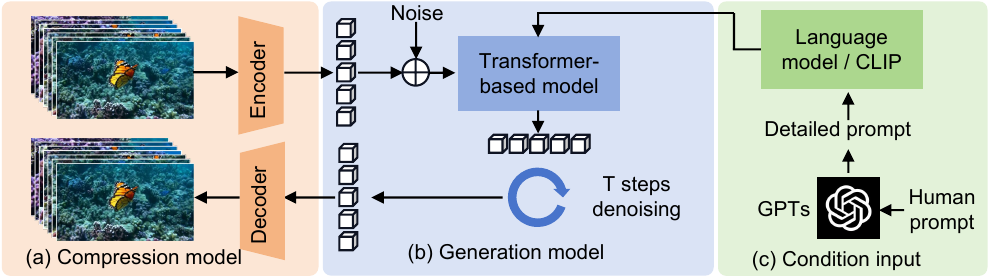}
  }

  \caption{An potential architecture of Sora. This architecture is inspired from \cite{videoworldsimulators2024, liu2024sora}. }
    \label{fig:sora}
\end{figure*}

\begin{figure*}[!t]
\centering

  \resizebox{1\linewidth}{!} {
    \includegraphics{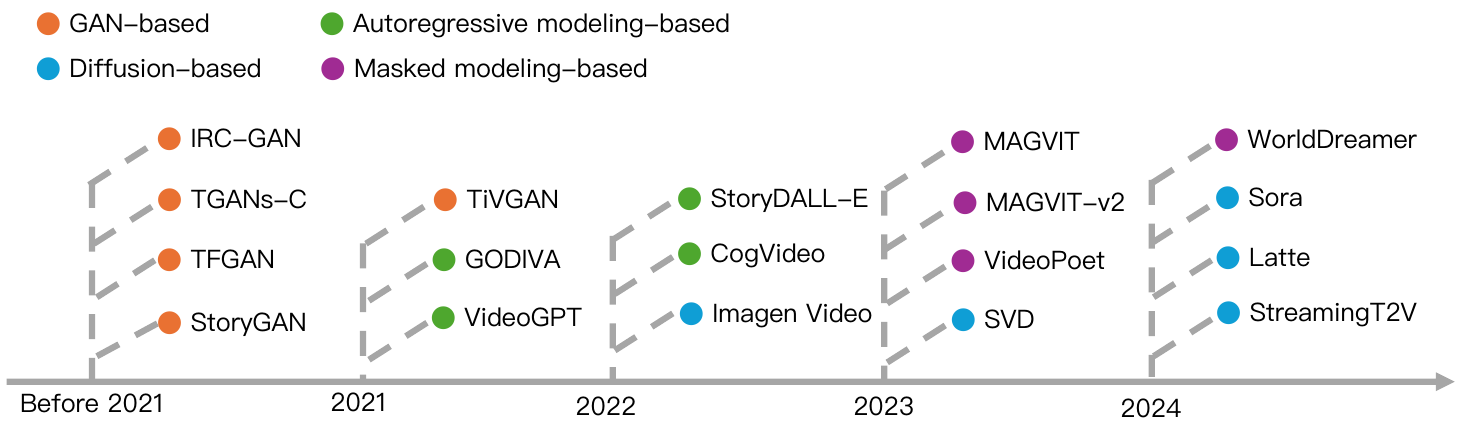}
  }

  \caption{Chronological overview of video generation models. We present representative models proposed in recent years. Before 2021, GAN-based models dominate video generation. After that, autoregressive modeling-based, diffusion-based, and masked modeling-based models start to emerge and achieve surprising performance.}
    \label{fig:dev_video_gen}
\end{figure*}

\noindent
\textbf{Masked Modeling.}
Masked modeling is first designed for self-supervised learning for language models \cite{devlin2018bert,liu2019roberta,jiao2019tinybert} and image models \cite{bao2021beit,he2022masked}. Given a sequence of tokens $\left(\mathbf{x}_1, \mathbf{x}_2, \ldots, \mathbf{x}_K\right)$, some tokens are randomly masked out.
Then, the model is forced to predict the masked tokens and reconstruct the original representation.  Noting the ability of image reconstruction of masked modeling, some works \cite{chang2022maskgit,li2023mage,li2023self} directly generate images from mask tokens and find it also generalizes well in video generation tasks \cite{yu2023magvit, yu2023language}. Considering its simplicity and surprising performance, it is also a promising direction for future generation techniques.

\subsection{Advanced Video Generation Models} \label{sec:adv_video}
In this section, we review the advanced video generation models proposed in recent years. Based on the given conditions (\eg example, classes, audios, texts, images, or videos), during generation, video generation tasks can be divided into different categories. Here, we mainly focus on the text-to-video method, where the text description is available during generation. These models aim to generate videos that are semantically aligned with given texts while maintaining consistency between different frames. The methods for idea generation with other conditions can be modified from the text-to-image models. 

\subsubsection{GAN-based Methods}
Besides the success of image generation, GAN-based models also achieve remarkable performance for video generation \cite{pan2017create,balaji2019conditional,li2022word,li2018video,li2019storygan,kim2020tivgan,deng2019irc}. Here, we select three representative methods and review them briefly.  We visualize a general architecture of GAN-based methods from video generation in Figure~\ref{fig:generation_methods} (a).

Temporal GANs conditioning on captions (TGANs-C) \cite{pan2017create} adopts a text encoder based on LSTM \cite{hochreiter1997long} to extract a text embedding.
This embedding is then combined with a vector of random noise, which together form the input to the generator. The generator contains a series of spatio-temporal convolutions to generate the frame sequence. Unlike the GANs-based models for image generation in Section~\ref{sec:geneartion_tech}, which typically has only one discriminator, TGANs-C designs three discriminators in video, frame, and motion-levels, respectively. Benefiting from these discriminators, the model is capable of producing videos that align with the provided text and akin to authentic video footage.


Text-Filter conditioning Generative Adversarial Network (TFGAN) \cite{balaji2019conditional} adopts the text features extracted from the text encoder to generate a series of filters in different frames. Then, these filters are employed as the convolutional filters in the discriminator for each frame generation. This operation enhances the semantic association between the given text and the generated video.

The StroyGAN \cite{li2019storygan} aims to generate a sequence of frames based on a multi-sentence paragraph, where each sentence is responsible for one frame. It adopts a story encoder and a context encoder to extract the global representation of the multi-sentence paragraph and sentence for the current frame, respectively. Then, the output from the story encoder and context encoder are combined and input to the generator to generate the current frame. It also employs two discriminators to ensure the frame-level and video-level consistency with the given paragraph.

\subsubsection{Diffusion-based Methods}
The development of diffusion models for image generation also facilitates the progress in video generation. We select four representative approaches due to their effectiveness or efficiency.
We summarize the framework of these methods in Figure~\ref{fig:generation_methods} (b).

Imagen Video \cite{ho2022imagen} proposes a cascaded sampling pipeline for video generation. Starting from a base video generation model \cite{ho2022video}, which generates video with low resolution and low frame rate, the authors cascade spatial and temporal super-resolution models to progressively improve the resolution and frame rate of generated videos.

Stable video diffusion (SVD) \cite{svd} is built upon Stable Diffusion \cite{rombach2022high} by inserting temporal convolution and attention layers after spatial convolution and attention blocks. To improve the generation performance, the authors propose to disengage the training into three stages: pre-training on text-to-image task, pre-training on text-to-video task, and text-to-video finetuning with high-quality data. It proves the importance of data curation for video diffusion models.

Latte \cite{ma2024latte} is an early attempt to apply a Transformer-based model in video generation. The model is built based on DiT \cite{peebles2023scalable} and contains extra blocks for spatial-temporal modeling. To ensure the efficiency in generation, the authors explore four efficient designs for spatial and temporal modeling, which is similar to the operations mentioned in Section~\ref{sec:visual_found}. The architecture of the Latte is thought to be similar to the design of Sora.

StreamingT2V \cite{henschel2024streamingt2v} divides the text-to-video generation into three steps, enabling to generation of long videos with even more than 1,200 frames. First, it employs pre-trained text-to-video models to generate a short video \eg with only 16 frames. Then, it extends a video diffusion model with short-term and long-term memory mechanisms to autoregressively generate further frames. Finally, another high-resolution video generation model is adopted to enhance generated videos.

\subsubsection{Autoregressive Modeling-based Methods}
Autoregressive modeling is also a popular technique in video generation \cite{yan2021videogpt, wu2021godiva, hong2022cogvideo, maharana2022storydall, yu2022scaling}
. We present its architecture in Figure~\ref{fig:generation_methods} (c).

VideoGPT \cite{yan2021videogpt} is a representative autoregressive modeling-based method. It first trains a VQ-VAE \cite{van2017neural} to encode videos into latent tokens. Then, the authors leverage a GPT-like framework \cite{radford2018improving} and train the model learning to predict the next token in the latent space. During the inference, a series of tokens is sampled from the latent space and the trained VideoGPT with VQ-VAE decodes it into generated videos.

GODIVA \cite{wu2021godiva} also generates videos in a similar way while emphasizing reducing the computation complexity of the model. Specifically, the authors propose to replace an original self-attention layer with three sparse self-attention layers, which only are conducted along the temporal, row, and column dimensions of the latent features, respectively. The effectiveness of this disentangling operation is also verified by models mentioned in Section~\ref{sec:tech_video}.

CogVideo \cite{hong2022cogvideo} inherits the knowledge from the pre-trained autoregressive model CogView2 \cite{ding2022cogview2} to reduce the burden of training from scratch. To improve the alignment between the given text and generated video, the authors propose a multi-frame-rate hierarchical generation framework, which first generates key frames in an autoregressive manner and then recursively interpolates frames with bidirectional attentions. 

\subsubsection{Masked Modeling-based Methods}
\begin{figure*}[!t]
\centering

  \resizebox{1\linewidth}{!} {
    \includegraphics{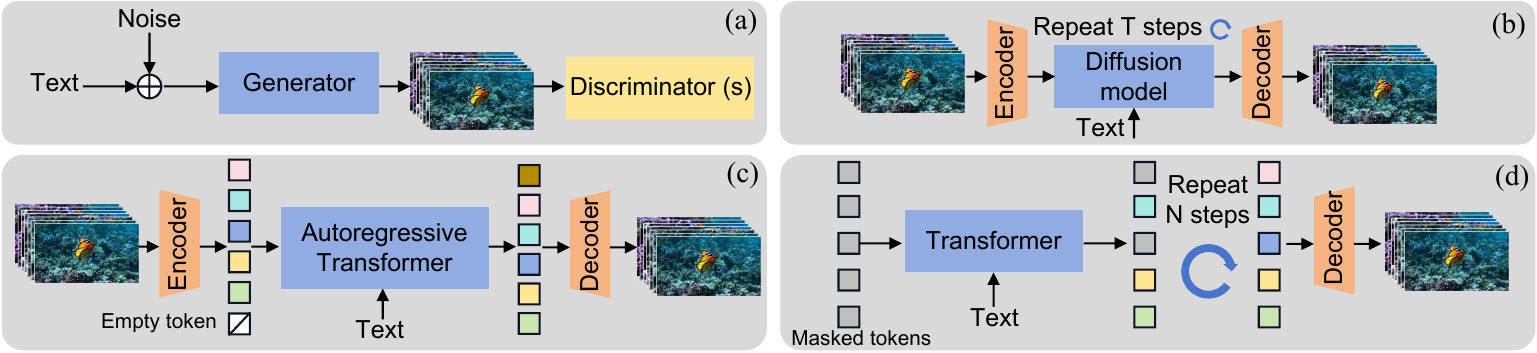}
  }

  \caption{Video generation methods. (a) GAN-based (b) Diffusion-based (c) Autoregressive modeling-based (d) Masked modeling-based.}
    \label{fig:generation_methods}

\end{figure*}

Masked modeling is also an emerging video generation method. Unlike autoregressive modeling, which suffers from sequential generation, the masked modeling method can decode videos in parallel. We visualize its architecture in Figure~\ref{fig:generation_methods} (d).

MAGVIT \cite{yu2023magvit} encodes videos into tokens through a 3D-VQ tokenizer and leverages a masked token modeling paradigm to accelerate the training. Specifically, the target tokens are randomly replaced with conditional tokens and masked tokens during training. Then, a bidirectional Transformer is trained to refine the conditional tokens, predict masked tokens, and reconstruct target tokens. To improve the generation quality, MAGVIT-v2 \cite{yu2023language} is introduced to improve the video tokenizer. The authors design a lookup-free quantization method to build the codebook and propose a joint image-video tokenization model, enabling it can tackle image and video generation jointly. After that, VideoPoet \cite{kondratyuk2023videopoet} integrates MAGVIT-v2 \cite{yu2023language} into a large language model to generate videos from various conditioning signals

Similarly, WorldDreamer \cite{wang2024worlddreamer} also trains to model to reconstruct masked tokens based on those unmasked tokens. To facilitate the training process, they design a spatial-temporal patchwise Transformer, which conducts attention within a spatial-temporal window. It adopts cross-attention layers to inject information of given text descripction into the model. The priority of parallel decoding enables it to achieve much faster video generation than diffusion-based and autoregressive-based methods.  

\subsubsection{Datasets and Evaluation Metrics}
Training a text-to-video generation model requires large-scale video-text pairs.  In Table~\ref{tab:video_dataset}, we present serveral popular datasets. These datasets may also be employed to train multi-modal models. Based on the technical report from Sora, the data quality, for example the video-text alginment and the richness of captions, is essential to the generation performance. Hence, we hope more large-scale high quality dataset can be open-sourced, prompting the prosperity of video generation and even the development of world models.

The  metrics adopted to evaulate the video generation performance varies in different papers. Fo example, Latte \cite{ma2024latte} and VideoGPT \cite{yan2021videogpt} measure the performance through Fréchet Video Distance (FVD) \cite{unterthiner2018towards}. CLIP similarity
(CLIPSim) \cite{wu2021godiva} is also a common evualtion approach. Human evaluation as complementary to these metrics is also widely adopted in existing works. Since evaluation score are highly related to the random seed, it is not easy to conduct fair comparison. Moreover, different mehtods may adopt differnt dataset to evalution performance, which further aggravates this problem. Human preference annotations may be a potential solution for video generation evaluation.
Recently, some comprehensive benchmarks \cite{liu2023evalcrafter, liu2024fetv, huang2023vbench} are proposed for the comparison fairness.

\subsection{Towards World Models: Sora} \label{sec:sora}
Sora is a closed-source text-to-video generation model developed by OpenAI. Besides being capable of generating a minute of high-fidelity video, it demonstrates some abilities to simulate the real world. It directs a way towards the world model through video generation models. In this section, we briefly introduce the techniques behind Sora. Since Sora is closed-source, all analyses here are mainly based on its technical report \cite{videoworldsimulators2024} and may vary from its real implementation.

\subsubsection{Framework}
Sora is thought to be a diffusion-based video generation model. It consists of three parts:
1. A compression model that compresses a raw video both temporally and spatially into latent representation and an asymmetrical model that maps the latent representation back to the original video. 2. A Transformer-based diffusion model, similar to DiT \cite{peebles2023scalable}, which is trained in the latent space. 3. A language model that encoders human instruction into embedding and injects it into the generation model.

\noindent
\textbf{Compression Model.}
The compression model usually contains an encoder and a decoder. The former is adopted to project the video into a low-dimensional latent space, while the latter maps the latent representation back to the video.  Based on the technical report \cite{videoworldsimulators2024}, the compression model is built based on VAE \cite{kingma2013auto} or VQ-VAE \cite{van2017neural}. Since the architecture of the decoder is usually in symmetric to the encoder, we mainly focus on the architecture of the encoder in this review.

Given a raw video $\mathbf{V} \in \mathbb{R}^{T \times H \times W \times C}$, the encoder first projects it into a sequence of tokens $\mathbf{x} \in \mathbb{R}^{n_t \times n_h \times n_w \times d}$. Based on the methods employed in visual foundation models mentioned in Section~\ref{sec:visual_found}, there exist two options: spatial-only compression and spatial-temporal compression. The spatial-only compression only compresses the video along the spatial dimension. It extracts image patches of size $h \times w$ for each frame and adopts a 2D convolutional layer to project it into $\mathbf{x}_i \in \mathbb{R}^{d}$. In this case, we have $n_t = T$, $n_h = H/h$, and $n_w = W/w$. This operation is widely adopted in ViTs \cite{dosovitskiy2020image}. The spatial-temporal compression method compresses the video along both the spatial and temporal dimensions, which provides a larger compression rate. Specifically, it extracts spatial-temporal tubes of size $t\times h \times w$ from the video and adopts a 3D convolutional layer to project it into an embedding $\mathbf{x}_i \in \mathbb{R}^{d}$. Thus, we have $n_t = T/t$,  $n_h = H/h$, and $n_w = W/w$. This operation is similar to the tubelet embedding technique in ViViT \cite{arnab2021vivit}.

After the tokenization, the encoder can further process these tokens through Transformer blocks, convolutional blocks, or the combination of them and project them into $z \in \mathbb{R}^{n_{t}^\prime \times n_{h}^\prime \times n_{w}^\prime \times d^\prime}$. We present the architecture of the compression model in Figure~\ref{fig:sora} (a).


\noindent
\textbf{Generation Model.}
Based on the technical report, the generation model is built up on DiT \cite{peebles2023scalable}. Since the original DiT is designed for class-to-image generation, two modifications should be conducted on it. First, since the self-attention blocks and MLP blocks in DiT are designed for spatial modeling, extra blocks for temporal modeling should be added. This could be achieved via extending the original self-attention to both spatial and temporal dimensions. Second, the condition is changed from class to text, and blocks to inject the text information should be added. The text-to-image cross-attention block is a potential solution, whose effectiveness has been proven in \cite{chen2023pixart}.
Based on this, one layer of the potential architecture can be formulated as:
\begin{equation}
    \mathbf{x}^\prime = \mathbf{x} + \operatorname{STA}(\mathbf{x}),
\end{equation}
\begin{equation}
    \mathbf{x}^{\prime \prime} = \mathbf{x}^\prime + \operatorname{CA}(\mathbf{x}^\prime, \mathbf{c}),
\end{equation}
\begin{equation}
    \mathbf{y} = \mathbf{x}^{\prime\prime} + \operatorname{MLP}(\mathbf{x}^{\prime\prime}),
\end{equation}
where $\operatorname{STA}$ and $\operatorname{CA}$ denotes the spatial-temporal attention and text-to-image cross attention blocks, respectively. $\mathbf{x}^g \in \mathbb{R}^{(n_t^{g} \times n_h^{g} \times n_w^{g}) \times d^{g}}$ denotes the input of this layer. The text embedding derived from a language model \eg  T5 \cite{raffel2020exploring} or a multi-modal model \eg CLIP \cite{radford2021learning} is denoted as $c$. We omit the injection of timestep information for brevity, which can be achieved with adaptive layer norm blocks \cite{perez2018film}. We also present the potential architecture in  Figure~\ref{fig:sora} (b). Finally, the generation model is trained to predict noise added to the latent representation $z$. More details can be found in diffusion techniques mentioned in Section~\ref{sec:diffusion}


\subsubsection{Training Data}
A large challenge to training Sora is collecting large-scale high-quality video-text pairs. Previous works \cite{betker2023improving, chen2023pixart} have proven that generation performance is highly dependent on the quality of data. Low-quality data, for example, noisy video-text pairs or too simple video captions, results in generation models with pool instruction following. To tackle this problem, Sora adopts the re-captioning technique proposed in DALL-E 3 \cite{betker2023improving}. Specifically,  a video captioner is trained with high-quality video-text pairs, where the text is well-aligned with the corresponding video and contains diverse and descriptive information. The video captioner could be a video version of multi-modal large language models, like GPT-4V \cite{achiam2023gpt}, mPLUG \cite{xu2023mplug}, or InternVideo \cite{wang2024internvideo2}. Then, the pre-trained video captioner is employed to generate high-quality captions for the training data of Sora. This simple method effectively improves the data quality.

During inference, to solve the problem that users may provide too simple prompts, Sora adopts GPT-4 \cite{achiam2023gpt} to rewrite the prompts so that they are detailed. This enables Sora to generate high-quality videos.






\subsubsection{Towards World Models}
Based on the claim from OpenAI, Sora can work as a world simulator, since it can understand the result of an action. For an example from its technical report, Sora generates a video where a painter can leave new strokes along a canvas that persist over time.  Another example is that a man can eat a burger and leave bite marks, which denotes that Sora can predict the results of eating. These two examples indicate that Sora can understand the world and predict the results of an action.  This capability is well-aligned with the target of world models: understanding the world via predicting the future. Hence, we believe that the techniques behind Sora can further inspire the exloration of world models.

First, the training and inference strategies improve the performance and efficiency in large generation models. For example, Sora learning from videos with enative aspect ratios, which obviously improve the composition and framing of generated videos. This requires both technical and engineering optimization to enbale efficent training. Generating videos with 1 minute length is a large chalenge and burden for inference server, which still impede releasing Sora to public until now. The OpenAI's soulution may be valubale for the community of large models. More potential techniues adopted in Sora can be found in \cite{liu2024sora}. We believe these contritbuons in Sora could also inpire building world models.

Second, Sora adopts Transformer-based generation with extensive parameters and large-scale training data, resuting in emergent abilities in video generation. This suggests that there also exsiting scaling laws in the visual field and directs a promising way to build large vision models or even world models.

Finally, Sora emphasize the essentiality of training data for good generation performance once again. Although OpenAI has not disclosed the sources and scale of data used in Sora, some guesses think extensive game videos may be introduced during training. The game videos may contains rich physical information, helping Sora to understanding the physical world. This indicates that the incorporating physical engine may be a potential path towards building world moels.


\begin{table*}[ht]

\renewcommand{\arraystretch}{1.0}
    \centering

    \vspace{1pt}
    \caption{Datasets for video generation. ASR: Automatic speech recognition. This table is reported by \cite{chen2024panda}}
\begin{tabular}{c | c | c | c | c | c | c | c | c}
\toprule
\textbf{Dataset} & \textbf{Year}  & \textbf{Text} & \textbf{Domain} & \textbf{$\#$Video} & \textbf{Avg} & \textbf{Video len} & \textbf{Avg text len} & \textbf{Resolution} \\ \midrule
MSVD \cite{chen2011collecting} & 2011 & Manual Caption  &  Open & 1970 & 9.7s & 5.3hr & 8.7 words & - \\ 
LSMDC \cite{rohrbach2015dataset} & 2015 & Manual Caption  &  Movie & 118K & 4.8s & 158hr & 7.0 words & 1080p \\ 
MSR-VTT \cite{xu2016msr} & 2016  & Manual Caption  &  Open & 10K & 15.0s & 40hr & 9.3 words & 240p \\ 
DiDeMo \cite{anne2017localizing} & 2017 & Manual Caption  &  Flickr & 27K & 6.9s & 87hr & 8.0 words & - \\ 
ActivityNet \cite{caba2015activitynet} & 2017 & Manual Caption & Action & 100K & 36.0s & 849hr & 13.5 words & - \\ 
YouCook2 \cite{zhou2018towards} & 2018 & Manual Caption  & Cooking & 14K & 19.6s & 176hr  & 8.8 words & - \\ 
VATEX \cite{wang2019vatex} & 2019 & Manual Caption & Open  & 41K &  ~10s & ~115hr & 15.2 words & - \\
HowTo100M \cite{miech2019howto100m} & 2019 & ASR & Open & 136M & 3.6s & 134.5Khr & 4.0 words & 240p \\ 
ACAV \cite{lee2021acav100m} & 2021 &  ASR &  Open & 100M &  10.0s &  277.7Khr & - & -  \\ 
YT-Temporal-180M \cite{zellers2021merlot} & 2021 &  ASR & Open  & 180M & - & - & - & - \\ 
HD-VILA-100M \cite{xue2022advancing} & 2021 &  ASR &  Open & 103M & 13.4s & 371.5Khr & 32.5 words & 720p  \\
WebVid-10M \cite{bain2021frozen} & 2021 & Manual Caption & Open & 10M & 18.0s & 50Khr & 12.0 words & - \\
Vimeo25M \cite{wang2023lavie} & 2023 & Automatic Caption & Open & 25M & 4.5s & - & 10.0 words & - \\
InternVid \cite{wang2023internvid} & 2023 & Automatic Caption & Open & 234M & 11.7s & 760.3Khr & 17.6 words & 720P \\
Panda-70M \cite{chen2024panda} & 2024 & Automatic Caption & Open & 70.8M & 8.5s & 166.8Khr & 13.2 words &  720p \\ 

\bottomrule
\end{tabular}

\label{tab:video_dataset}
\end{table*}

\section{World Models for Autonomous Driving}

\begin{figure*}[t]
	\centering
	\includegraphics[width=1\textwidth]{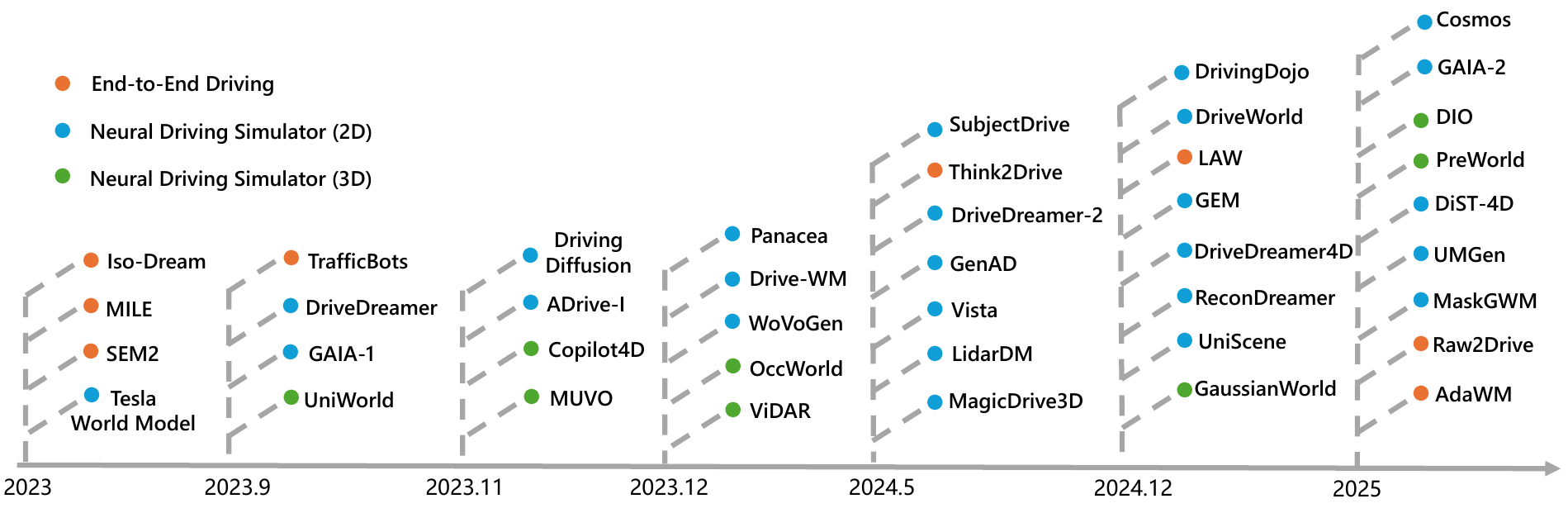} 
	\caption{Timeline of World Models in Autonomous Driving. End-to-end driving and neural driving simulator (both 2D and 3D) approaches are emerging since 2023.}
	\label{fig:wm_ad}
\end{figure*}

Driving requires navigating uncertainty. It is crucial to understand the uncertainty inherent in autonomous driving to make safe decisions, where even a minor mistake could have fatal consequences~\cite{hu}. There are two primary forms of uncertainty: epistemic uncertainty, which stems from a deficit in knowledge or information, and aleatoric uncertainty, which is rooted in the inherent randomness of the real world~\cite{fox2011distinguishing}. To ensure safe driving, it is imperative to leverage past experiences embedded in world models to effectively mitigate both aleatoric and epistemic uncertainty.

World models are adept at representing an agent's spatio-temporal knowledge about its environment through the prediction of future changes~\cite{lecun2022jepa}. Two primary types of world models exist within autonomous driving aimed at reducing driving uncertainty, i.e., world model for end-to-end driving and world model as neural driving simulator. In the simulation environment, methods such as MILE~\cite{mile} and TrafficBots~\cite{trafficbots} do not distinguish between epistemic and aleatoric uncertainties and incorporate them into the model based on reinforcement learning, enhancing their capacity for decision-making and future prediction, thereby paving the way to end-to-end autonomous driving. In the real environment, Tesla~\cite{tesla} and methods like GAIA-1~\cite{gaia} and Copilot4D~\cite{copilot4d} involve utilizing generative models to construct neural driving simulators that produce 2D or 3D future scenes to enhance predictive capabilities, thus reducing aleatoric uncertainty. Additionally, generating new samples can mitigate epistemic uncertainty regarding rare instances such as corner cases. Figure~\ref{fig:wm_ad} illustrates these two types of world models in autonomous driving. The neural driving simulator can be further subdivided into two categories: those generating 2D images and those simulating 3D scenes.

\subsection{End-to-end Driving}


\begin{table*}[t]
	\centering
	\caption{Summary of end-to-end driving methods based on world models. {Img}, {Act}, {Seg}, and {Dest} stand for images, action, segmentation, and destination, respectively.}
	\resizebox{1.0\textwidth}{!}
	{
		\begin{tabular}{c|c|c|c|c|c|c}
			\toprule
			\textbf{Method}&\textbf{Type}&\textbf{Core Structure}&\textbf{Reward}&\textbf{Input}&\textbf{Output}&\textbf{Simulator}\\
			\midrule
			Iso-Dream~\cite{isodream}&Reinforcement Learning&RSSM~\cite{hafner2019planet} &$\checkmark$&Img, Act &Img, Act&CARLA v1\\
			MILE~\cite{mile}&Imitation Learning& PGM~\cite{kingma2013auto}&$\times$&{Img}, {Act}&{Img}, {Act}, BEV {Seg}&CARLA v1\\
			SEM2~\cite{sem2}&Reinforcement Learning&RSSM~\cite{hafner2019planet}&$\checkmark$&Img, Act&Img, Mask&CARLA v1\\
			TrafficBots~\cite{trafficbots}&Reinforcement Learning&CVAE~\cite{sohn2015learning}&$\checkmark$&Static Map, Traffic Lights, Act &Act, Dest&CARLA v1\\
			Think2Drive~\cite{think2drive}&Reinforcement Learning&RSSM~\cite{hafner2019planet}&$\checkmark$&Box, HD-Map, Traffic Lights&Act&CARLA v2\\
			\bottomrule
		\end{tabular}
	}
	\label{tab:e2e_structure}
\end{table*}

In the domain of autonomous driving, the development of world models assumes a crucial role as they strive to construct dynamic representations of environments. Accurate predictions about the future are imperative for ensuring safe maneuvering in contexts. However, constructing world models for autonomous driving poses distinct challenges, mainly originating from the intricate sample complexity in driving scenarios. end-to-end autonomous driving methods~\cite{isodream,mile,sem2} strive to tackle these challenges by minimizing the search space and integrating explicit disentanglement of visual dynamics on the CARLA simulator~\cite{carla}. The comparison of existing end-to-end driving methods based on world models is illustrated in Table~\ref{tab:e2e_structure}.

Iso-Dream~\cite{isodream} introduces a Model-Based Reinforcement Learning (MBRL) framework, aimed at effectively disentangling and utilizing controllable and noncontrollable state transitions via reinforcement learning. Furthermore, Iso-Dream optimizes the agent's behavior based on the separated latent imaginations of world models. In detail, Iso-Dream projects non-controllable states into the future to estimate state values and links them with the current controllable state. Iso-Dream enhances the agent's long-horizon decision-making capabilities, exemplified in scenarios like autonomous vehicles proactively evade potential hazards by anticipating the movements of surrounding vehicles.

Iso-Dream learns the world model by mapping the 2D image in front view to control signals, which is not suitable for autonomous driving in 3D space. To address this issue, MILE~\cite{mile} integrates the world model with imitation learning in 3D space, i.e., Bird's Eye View (BEV) space. MILE uses 3D geometry as an inductive bias and creates a latent space from expert driving videos. The training occurs using an offline dataset of urban driving, devoid of any necessity for online engagement with the scene. In performance, it surpasses prior cutting-edge methods by a significant 31\% margin in driving score on CARLA, even when operating in entirely new town and weather conditions. Moreover, MILE demonstrates its capability to execute intricate driving maneuvers solely based on plans generated through imaginative processes.

Similar to MILE, SEM2~\cite{sem2} also constructs a world model in 3D space. SEM2 employs a novel approach by incorporating a latent filter to isolate crucial task-specific features and then utilizes these features to reconstruct a semantic mask. Additionally, it utilizes a multi-source sampler during training, which merges standard data with various corner case data within a single batch, effectively ensuring a balanced data distribution. Specifically, SEM2 takes camera and LiDAR as inputs, encoding them into a latent state with deterministic and stochastic variables. The initial latent state is subsequently employed to regenerate the observation. Following this, the latent semantic filter isolates driving-relevant features from the latent state, reconstructs the semantic mask, and predicts the reward. Extensive experiment conducted on the CARLA simulator showcases SEM2's adeptness in sample efficiency and robustness to variations in input permutations.

TrafficBots~\cite{trafficbots}, another end-to-end driving method based on world model, places emphasis on forecasting the actions of individual agents within a given scenario. By factoring in the destination of each agent, TrafficBots utilizes a Conditional Variational Autoencoder (CVAE)~\cite{sohn2015learning} to imbue individual agents with unique characteristics, enabling action anticipation from a BEV perspective. TrafficBots offers quicker operational speeds and scalability to handle larger numbers of agents. Experiments carried out on the Waymo dataset illustrate TrafficBots' capacity to emulate realistic multi-agent behaviors and attain promising results in motion prediction tasks.

The above methods~\cite{isodream,mile,sem2,trafficbots} were experimented in CARLA v1, but inherently face challenges regarding data inefficiency in CARLA v2. CARLA v2 offers a more quasi-realistic testbed. Addressing the complexities of CARLA v2 scenarios, Think2Drive~\cite{think2drive}, a model-based reinforcement learning method for autonomous driving, encourages the planner to \textit{think} within the learned latent space. This approach significantly enhances training efficiency by utilizing a low-dimensional state space and leveraging parallel computing of tensors. Think2Drive achieves expert-level proficiency on CARLA v2 simulator after a mere 3-day training period utilizing a single A6000 GPU.
Furthermore, Think2Drive introduces the CornerCase Repository, a novel benchmark designed to assess driving models across diverse scenarios. 

Despite the advancements seen in world models for end-to-end driving using reinforcement learning, a significant limitation remains: its primary emphasis on simulation environments. Next, we will delve into research on world models for autonomous driving in real-world scenarios.

\subsection{Neural Driving Simulator}
High-quality data serves as the bedrock for training deep learning models. While text and image data are readily available at low costs, acquiring data in the realm of autonomous driving poses challenges owing to factors such as spatio-temporal complexities and concerns regarding privacy. This is particularly true for addressing long-tail targets that directly impact realistic driving safety. World models are pivotal for understanding and simulating the complex physical world~\cite{gaia}. Some recent endeavors have introduced diffusion models~\cite{ho2020denoising} into the domain of autonomous driving to build world models as neural simulators to generate requisite autonomous 2D driving videos~\cite{gaia,drivedreamer,genad,subjectdrive}. Additionally, some methods employ world models to generate 3D occupancy grids or LiDAR point clouds depicting future scenes~\cite{copilot4d,muvo,vidar,driveworld}. Table~\ref{tab:NDS_ad_structure} provides an overview of these neural driving simulator methods based on world models.

\subsubsection{2D Scene Generation}
World models for driving video generation entail tackling two pivotal challenges: \textit{Consistency} and \textit{Controllability}. Consistency is crucial for maintaining temporal and cross-view coherence between generated images, whereas controllability ensures that generated images align with corresponding annotations~\cite{wen2023panacea}. The comparison of exiting 2D driving video generation methods based on world models are shown in Table~\ref{tab:videoGen}.

\begin{table*}[t]
	\centering
	\caption{Summary of model structure for neural driving simulator based on world models in autonomous driving. {Img}, {Act}, {PC}, {Traj}, {Occ}, {HD}, {Flow}, {Lay}, {Obj}, {Seq}, and {Arg2} stand for images, action, point cloud, trajectory, occupancy, HD-Map, optical flow, layout, objects, sequence, and Argoverse2, respectively.}
	\resizebox{1.0\textwidth}{!}
	{
		\begin{tabular}{c|c|c|c|c|c|c|c}
			\toprule
			\textbf{Task}&\textbf{Method}&\textbf{Data Source}&\textbf{Architecture}&\textbf{Encoder}&\textbf{Decoder}&\textbf{Input}&\textbf{Output}\\
			\midrule
			\multirow{10}*{{\makecell[c]{2D}}}&GAIA-1~\cite{gaia}&Wayve~\cite{gaia}& GPT& VQ-VAE& Video Diffusion Decoder&Img, Text, Act&Img\\
			&DriveDreamer~\cite{drivedreamer}&nuScenes~\cite{nuscenes}&Diffusion&VAE&Task Specific Decoder&Img, Act, Box, Text&Img, Act\\
			&DrivingDiffusion~\cite{drivingdiffusion}&nuScenes~\cite{nuscenes}&Diffusion&Diffusion Encoder&Diffusion Decoder&Img, Flow, Text, 3D Lay&Img\\
			&ADriver-I~\cite{adriver-i}&
			nuScenes~\cite{nuscenes}&Diffusion&CLIP-ViT&Video Diffusion Decoder&Img, Act&Img, Act \\
			&Panacea~\cite{wen2023panacea}&nuScenes~\cite{nuscenes}&Diffusion&Diffusion Encoder&Diffusion Decoder&Img, Text, BEV Seq &Img\\
			&Drive-WM~\cite{drivewm}&nuScenes~\cite{nuscenes}&Diffusion&VAE&VAE Decoder&Img, Act&Img, Traj\\
			&WoVoGen~\cite{wovogen}&nuScenes~\cite{nuscenes}&Diffusion&4D Volume Encoder&Diffusion Decoder&Img, Text, HD, Occ, Obj&Img, HD, Occ\\
			&DriveDreamer-2~\cite{drivedreamer2}&nuScenes~\cite{nuscenes}&Diffusion&VAE&Video Decoder&Img, HD, Traj, Box, Text&Img \\
			&GenAD~\cite{genad}&OpenDV-2K~\cite{genad}, nuScenes~\cite{nuscenes}&Diffusion&Diffusion Encoder&Diffusion Decoder&Img, Text, Act&Img\\	
			&SubjectDrive~\cite{subjectdrive}&nuScenes~\cite{nuscenes}&Diffusion&Diffusion Encoder&Diffusion Decoder&Img, Subject&Img\\
			\midrule
			\multirow{7}*{\makecell[c]{3D} }&UniWorld~\cite{uniworld}&nuScenes~\cite{nuscenes}&Transformer&BEV Encoder&Task Specific Decoder&Img&Img, Occ\\
			&Copilot4D~\cite{copilot4d}&nuScenes~\cite{nuscenes}, KITTI~\cite{kitti}, Arg2~\cite{argoverse2}&Diffusion& VQ-VAE&VQ-VAE Decoder&PC, Act&PC\\
			&MUVO~\cite{muvo}&CARLA v1~\cite{carla}&GRU&SensorFusion&Task Specific Decoder&Img, PC, Act&Img, Act, Occ\\
			&OccWorld~\cite{occworld}&nuScenes~\cite{nuscenes}&GPT& VQ-VAE&VQ-VAE Decoder&Occ, Ego Poses&Occ, Ego Poses\\
			&ViDAR~\cite{vidar}&nuScenes~\cite{nuscenes}&Transformer&BEV Encoder&Latent Render&Img, Ego Poses&PC\\
			&LidarDM~\cite{lidardm}& KITTI-360~\cite{kitti360}, Waymo~\cite{waymo}&Diffusion&Diffusion Encoder&Diffusion Decoder&Img, Traffic Lay&PC\\
			&DriveWorld~\cite{driveworld}&nuScenes~\cite{nuscenes}, OpenScene~\cite{openscene}&Transformer&BEV Encoder&Task Specific Decoder&Img, Act, Ego Poses&Occ, Act\\
			\bottomrule
		\end{tabular}
	}
	\label{tab:NDS_ad_structure}
\end{table*}

\begin{table}[t]
	\centering
	\caption{Comparison of FVD and FID metrics with 2D driving video generation methods based on world models on the validation set of the nuScenes dataset.}
	\resizebox{0.48\textwidth}{!}
	{
		\begin{tabular}{c|c|c|c|c}
			\toprule
			\textbf{Method}&\textbf{Multi-View}&\textbf{Multi-Frame}&\textbf{FVD}$\downarrow$&\textbf{FID}$\downarrow$\\
			\midrule
			DriveDreamer~\cite{drivedreamer}&&$\checkmark$&452.0&52.6 \\
			DriveDreamer~\cite{drivedreamer}&$\checkmark$&$\checkmark$&340.8&14.9 \\ 
			ADriver-I~\cite{adriver-i}&&$\checkmark$&97.0&5.5 \\
			WoVoGen~\cite{wovogen} &$\checkmark$&$\checkmark$&418.0&27.6 \\
			DrivingDiffusion~\cite{drivingdiffusion}&$\checkmark$&$\checkmark$&332.0&15.8 \\
			Panacea~\cite{wen2023panacea}&$\checkmark$&$\checkmark$&139.0&17.0 \\
			SubjectDrive~\cite{subjectdrive}&$\checkmark$&$\checkmark$&124.0&16.0 \\
			GenAD-nus~\cite{genad}&$\checkmark$&$\checkmark$&244.0&15.4 \\ 
			GenAD-OpenDV~\cite{genad}&$\checkmark$&$\checkmark$&184.0&15.4 \\ 
			Drive-WM~\cite{drivewm}&$\checkmark$&$\checkmark$&122.7&15.8 \\ 
			DriveDreamer-2~\cite{drivedreamer2}&$\checkmark$&$\checkmark$&55.7&11.2 \\ 
			\bottomrule
		\end{tabular}
	}
	\label{tab:videoGen}
\end{table}

GAIA-1~\cite{gaia} is a cutting-edge generative world model designed to produce lifelike driving videos, offering precise manipulation of both ego-vehicle actions and environmental elements. GAIA-1 tackles the challenge of world modeling by leveraging video, text, and action inputs as sequences of tokens, predicting subsequent tokens in an unsupervised way. Its structure comprises two main elements: the world model and the video diffusion decoder. The world model, boasting 6.5 billion parameters, underwent a 15-day training period utilizing 64 NVIDIA A100s, while the video decoder, with 2.6 billion parameters, was trained for the same duration using 32 NVIDIA A100s. The world model meticulously examines the elements and dynamics within the scene, whereas the diffusion decoder transforms latent representations into high-fidelity videos imbued with intricate realism. GAIA-1's training corpus comprises 4,700 hours of driving videos collected in London, spanning from 2019 to 2023. Notably, GAIA-1 demonstrates an understanding of 3D geometry and can capture the complex interactions induced by road irregularities. Furthermore, GAIA-1 adheres to similar scaling laws observed in Large Language Models (LLMs). With its learned representations and control over scene elements, GAIA-1 opens new possibilities for enhancing embodied intelligence.

While GAIA-1 can generate realistic autonomous driving scene videos, its controllability is limited to using only text and action as conditions for video generation, whereas autonomous driving tasks require adherence to structured traffic constraints. DriveDreamer~\cite{drivedreamer}, which excels in controllable driving video generation, seamlessly aligns with text prompts and structured traffic constraints, including HD-Map and 3D box data. The training pipeline of DriveDreamer comprises two stages: initially, DriveDreamer is trained with traffic structural information as intermediate conditions, significantly improving sampling efficiency. In the subsequent stage, the world model is developed through video prediction, where driving actions are iteratively utilized to update future traffic structural conditions. This enables DriveDreamer to anticipate variations in the driving environment based on different driving strategies. Through extensive experiments on the challenging nuScenes~\cite{nuscenes} benchmark, DriveDreamer is confirmed to enable precise and controllable video generation, representing the structural constraints of real-world traffic situations.

To further bolster the consistency and controllability of generated multi-view videos, DriveDreamer-2 \cite{drivedreamer2} is introduced as an evolution of the DriveDreamer framework. DriveDreamer-2 integrates a LLM to augment the controllability of video generation. Initially, DriveDreamer-2 integrates an LLM interface to interpret user queries and translate them into agent trajectories.  Subsequently, it generates an HD-Map in accordance with traffic regulations based on these trajectories. Additionally, DriveDreamer-2 proposes the unified multi-miew model to improve temporal and spatial consistency to generate multi-view videos. 

Different from DriveDreamer-2 with LLM, ADriver-I~\cite{adriver-i} leverages Multimodal Large Language Models (MLLMs) to enhance the controllability of generating driving scene videos. Inspired by the interleaved document approach in MLLMs, ADriver-I introduces interleaved vision-action pairs to establish a standardized format for visual features and their associated control signals. These vision-action pairs are utilized as inputs, and ADriver-I forecasts the control signal of the present frame in an autoregressive manner. ADriver-I continues this iterative process with the predicted next frame, enabling it to achieve autonomous driving in the synthesized environment. Its performance is rigorously assessed through extensive experimentation on datasets such as nuScenes~\cite{nuscenes} and sizable proprietary datasets.

ADriver-I is limited to generating single-view videos. To generate multi-view videos as DriveDreamer-2, Panacea~\cite{wen2023panacea} and DrivingDiffusion~\cite{drivingdiffusion} are proposed. Panacea~\cite{wen2023panacea} is an innovative video generation system designed specifically for panoramic and controllable driving scene synthesis. It operates in two stages: initially crafting realistic multi-view driving scene images, then expanding these images along the temporal axis to create video sequences.
For panoramic video generation, Panacea introduces decomposed 4D attention, enhancing both multi-view and temporal coherence. Additionally, Panacea utilizes ControlNet to incorporate BEV sequences. Beyond these fundamental features, Panacea maintains flexibility by enabling manipulation of global scene attributes through textual descriptions, including weather, time, and scene details, providing a user-friendly interface for generating specific samples. DrivingDiffusion~\cite{drivingdiffusion} also presents a multi-stage approach for generating multi-view videos. It involves several crucial stages: multi-view single-frame image generation, shared single-view video generation across multiple cameras, and post-processing capable of handling extended video generation. It also introduces local prompts to improve the quality of images effectively. Subsequent to the generation process, post-processing is employed to enhance the coherence among different views in subsequent frames. Additionally, it utilizes a temporal sliding window algorithm to prolong the video duration.

The objective of the above methods is to generate realistic driving scenario videos given certain conditions. Drive-WM~\cite{drivewm} takes this a step further by utilizing predicted future scene videos for end-to-end planning applications to enhance driving safety. Drive-WM introduces multi-view and temporal modeling to generate multi-view frames. To improve multi-view consistency, Drive-WM proposes factorizing the joint modeling to predict intermediate views conditioned on adjacent views, significantly enhancing consistency between views. Drive-WM also introduces a simple yet effective unified condition interface, enabling flexible utilization of diverse conditions such as images, text, 3D layouts, and actions, thereby simplifying conditional generation.
Furthermore, by leveraging the multi-view world model, Drive-WM explores end-to-end planning applications to enhance autonomous driving safety. Specifically, at each time step, Drive-WM utilizes the world model to generate predicted future scenarios for trajectory candidates sampled from the planner. These futures are evaluated using an image-based reward function, and the optimal trajectory is selected to extend the planning tree. Testing on real-world driving datasets validates Drive-WM's capability to produce top-tier, cohesive, and manageable multi-view driving videos, thereby unlocking avenues for real-world simulations and safe planning.

Control signals like bounding boxes or HD-Maps provide a sparse representation of the driving scene. WoVoGen~\cite{wovogen} enhances diffusion-based generative models by introducing a 4D world volume. Initially, WoVoGen builds a 4D world volume by merging a reference scene with a forthcoming vehicle control sequence. This volume then guides the generation of multi-view imagery. Within this 4D structure, each voxel is enriched with LiDAR semantic labels obtained via the fusion of multi-frame point clouds, enhancing the depth and complexity of environmental comprehension.

SubjectDrive~\cite{subjectdrive} has undertaken further research to explore the effects of increasing the scale of generated videos on the performance of perception models in autonomous driving. Through their investigations, they have demonstrated the efficacy of scaling generative data production in continuously enhancing autonomous driving applications. It has pinpointed the pivotal significance of enhancing data diversity in efficiently expanding generative data production. Consequently, SubjectDrive has developed an innovative model incorporating a subject control mechanism.

The above methods for generating driving videos have largely been studied on relatively small datasets like nuScenes~\cite{nuscenes}. GAIA-1~\cite{gaia} was trained on a dataset of 4,700 hours, but the training dataset is not publicly available. Recently, GenAD~\cite{genad} has released the largest multimodal video dataset for autonomous driving, OpenDV-2K, exceeding the scale of the widely used nuScenes dataset by a multiplier of 374. OpenDV-2K contains 2,059 hours of video content accompanied by textual annotations, drawn from a combination of 1,747 hours sourced from YouTube and an additional 312 hours gathered from public datasets. Addressing common challenges such as causal confusion and handling large motions, GenAD utilizes causal temporal attention and decoupled spatial attention mechanisms to effectively capture the rapid spatio-temporal fluctuations present in highly dynamic driving environments. This architecture allows GenAD to generalize across diverse scenarios in a zero-shot way. This acquired understanding is further substantiated through the application of its learned knowledge to driving challenges, including planning and simulation tasks.

\subsubsection{3D Scene Generation}
In addition to generating 2D videos for autonomous driving through world modeling, some methods delve into utilizing world models to produce 3D LiDAR point clouds or 3D occupancy grids. 

Copilot4D~\cite{copilot4d} presents an innovative approach to world modeling by first tokenizing LiDAR point cloud observations with VQ-VAE~\cite{van2017neural}, then predicting future LiDAR point clouds via discrete diffusion. To efficiently decode and denoise tokens in parallel, Copilot4D modifies the masked generative image Transformer to fit within the discrete diffusion framework with slight adjustments, yielding significant improvements. When utilized for training world models based on LiDAR point cloud observations, Copilot4D achieves a remarkable reduction of over 65\% in Chamfer distance for point cloud forecasting at 1s prediction and over 50\% at 3s prediction across datasets such as nuScenes~\cite{nuscenes}, Argoverse2~\cite{argoverse2}, and KITTI Odometry~\cite{kitti}. 

Copilot4D utilizes unannotated LiDAR data to construct its world model, while OccWorld~\cite{occworld} delves into the 3D occupancy space for the representation of 3D scenes. OccWorld initiates its approach by employing a VQ-VAE~\cite{van2017neural} to refine high-level concepts and derive discrete 3D semantic occupancy scene tokens in a self-supervised manner. Subsequently, it customizes the GPT~\cite{brown2020language} architecture, introducing a spatial-temporal generative Transformer to forecast scene tokens and ego tokens. Through these advancements, OccWorld achieves significant results in 4D occupancy forecasting and planning.

Copilot4D and OccWorld employ past LiDAR or 3D occupancy frames to generate future 3D scenes, whereas MUVO~\cite{muvo} adopts a more comprehensive strategy by leveraging raw camera and LiDAR data as input. MUVO aims to acquire a sensor-agnostic geometric representation of the environment and predicts future scenes in the forms of RGB images, 3D occupancy grids, and LiDAR point clouds.
Initially, MUVO undertakes image and LiDAR point cloud processing, encoding, and fusion utilizing a Transformer-based architecture. Subsequently, it inputs the latent representations of the sensor data into a transition model to establish a probabilistic model of the current state. Concurrently, MUVO forecasts the probabilistic model of future states and generates samples from it.

While Copilot4D, OccWorld, and MUVO generate 3D scenes without control, LidarDM~\cite{lidardm} excels in producing layout-aware LiDAR videos. LidarDM employs latent diffusion models to generate the 3D scene, integrating dynamic actors to establish the underlying 4D world, and subsequently generating realistic sensory observations within this virtual environment. Beginning with the input traffic layout at time $t = 0$, LidarDM initiates the generation process by creating actors and the static scene. Subsequently, LidarDM generates the motion of the actors and the ego-car, composing the underlying 4D world. Finally, a generative- and physics-based simulation is utilized to produce realistic 4D sensor data.
The LiDAR videos generated by LidarDM are realistic, layout-aware, physically plausible, and temporally coherent. They demonstrate a minimal domain gap when tested with perception modules trained on real data.

As an abstract spatio-temporal representation of reality, the world model possesses the capability to predict future states based on the present. The training mechanism of world models holds promise in establishing a foundational pre-trained model for autonomous driving. UniWorld~\cite{uniworld}, ViDAR~\cite{vidar}, and DriveWorld~\cite{driveworld} delve into the exploration of 4D pre-training based on world models, aiming to enhance various downstream tasks of autonomous driving, such as perception, prediction, and planning.

UniWorld~\cite{uniworld} introduces the concept of predicting future 3D occupancy as a pre-text task for autonomous driving, leveraging extensive unlabeled image-LiDAR pairs for 4D pre-training. It takes multi-view images as inputs, generating feature maps in a unified BEV space~\cite{sts}. These BEV representations are then utilized by a world model head to predict the occupancy of future frames. UniWorld demonstrates improvements in intersection over union for tasks like semantic scene completion and motion prediction compared to 3D pre-training methods~\cite{uniscene,min2023occupancy}.

While UniWorld has demonstrated the effectiveness of 4D pre-training based on world models for autonomous driving, it predicts future scenes by adding a simple occupancy head. ViDAR~\cite{vidar} proposes latent rendering
operator with differentiable ray-casting for future scene prediction.
ViDAR consists of three main components: history encoder, latent rendering operator, and future decoder. The history encoder embeds visual sequences into BEV space. Subsequently, these BEV features undergo processing by the latent rendering operator, which significantly bolsters downstream performance. The future decoder, functioning as an autoregressive Transformer, utilizes historical BEV features to iteratively forecast future LiDAR point clouds for various timestamps.

To enhance 4D pre-training for autonomous driving by better capturing spatio-temporal dynamics, DriveWorld~\cite{driveworld} takes a further step by separately addressing temporal and spatial information. DriveWorld introduces the memory state-space model to reduce uncertainty within autonomous driving across both spatial and temporal dimensions.
Firstly, to tackle aleatoric uncertainty, DriveWorld proposes the dynamic memory bank module, which learns temporal-aware latent dynamics to predict future scenes. Secondly, to mitigate epistemic uncertainty, DriveWorld introduces the static scene propagation module, which learns spatial-aware latent statics to provide comprehensive scene context.
Moreover, DriveWorld introduces the task prompt, utilizing semantic cues as guidance to dynamically adjust the feature extraction process for various driving tasks.

\section{World Models for Autonomous Agents}

\begin{figure*}[t]
  \centering
  \includegraphics[width=0.99\textwidth]{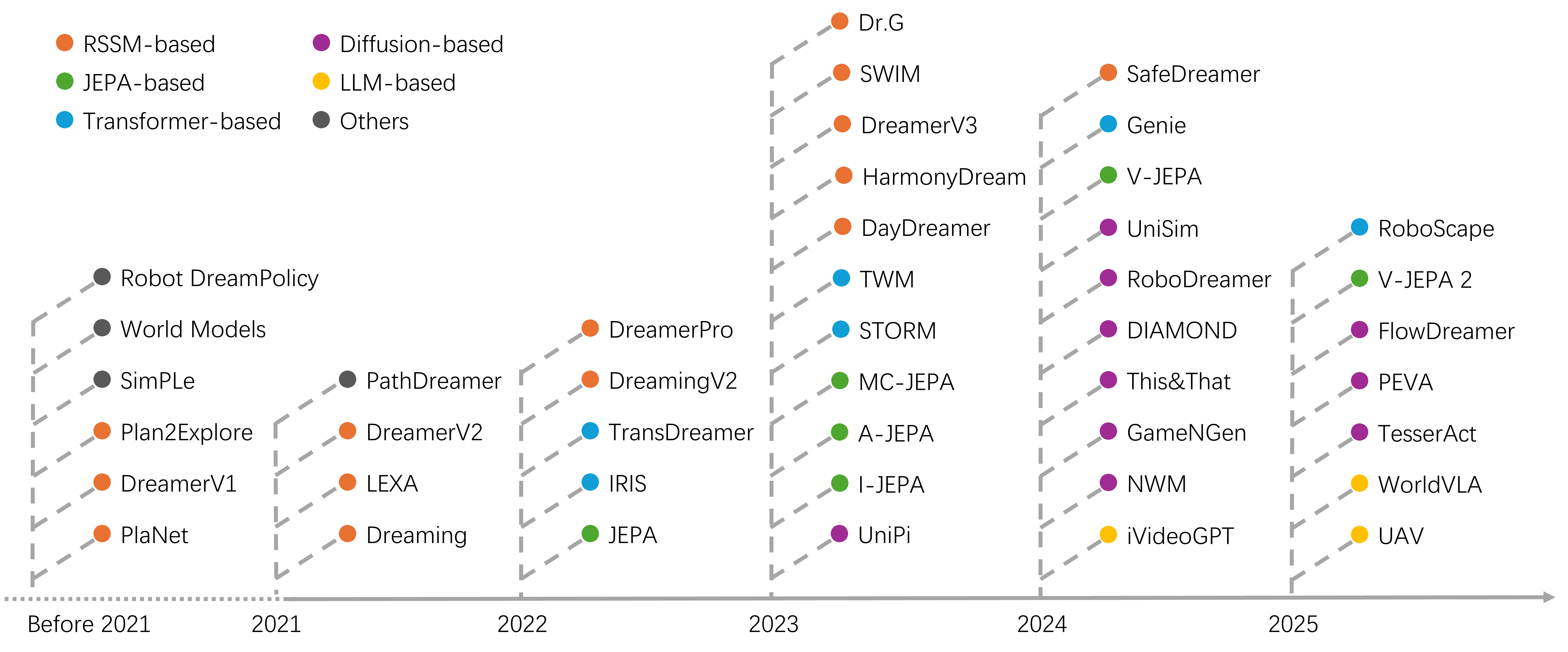} 
  \caption{Chronological overview of video generation model. We present the world model-based autonomous agents proposed in recent years. The colors show different structures of world models. The RSSM dominated these efforts while the Transformer, JEPA, and diffusion are gaining more and more attention from 2022.}
  \label{fig:wm_ag}
\end{figure*}

In artificial intelligence, an autonomous agent refers to a system that can perceive its surrounding environment through sensors (such as cameras) and act upon it through actuators to achieve specific goals\cite{franklin1997agent}. These agents can be physical, like robots, or virtual, such as software programs that perform tasks in digital environments.

Given a goal, agents need to plan a sequence of actions. There are already many successful algorithms for dynamic planning in known environments. In most cases, however, the environment is complex and stochastic, making it difficult to model by human experience explicitly. Therefore, this field's core topic is how agents learn to plan in an unknown and complex environment. One way to solve this problem is to have the agent accumulate experience and learn behaviors directly from the interaction with the environment, without modeling the state changes of the environment (the so-called model-free reinforcement learning). While this solution is simple and flexible, the learning process relies on many interactions with the environment, which may be extremely expensive, even unacceptable. 

World Models\cite{ha2018worldmodel} is the first work that introduces the concept of the world model in the field of reinforcement learning, modeling knowledge about the world from the agent's experience and gaining the ability to predict the future. This work demonstrates that even a simple RNN model can capture the dynamics of the environment and support the agent to learn and evolve policies in this model. This learning paradigm is referred to as \textit{learning in imagination} \cite{hafner2020dreamerV1}. With world models, the cost of trials and failures can be greatly reduced\cite{wu2023daydreamer}.

In this section, we introduce the world models for autonomous agents. We first describe the general framework of a world model-based agent, including the key components and the model structures widely used in world model-based agents in Section \ref{sec:ag_framework}. Then, we introduce the agents serving a variety of tasks, such as game agents and robotics, in Section \ref{sec:ag_task}. Finally, we present the benchmarks that are commonly used to evaluate the performance of world model-based agents.

\subsection{General Framework of an Agent based on World Model}\label{sec:ag_framework}
Most works implement world model-based agents under a basic framework originating from robotics. In the framework, the world model is the core component. To model and predict the surrounding environment, pioneers proposed several effective structures, which are widely used in later works. In this section, we describe in detail the key components of the framework and the widely used structures of world models.

\subsubsection{Key Components}
From the view of software engineering, an agent system can be decomposed into four components\cite{sakagami2023robotic}:

\noindent \textbf{Sensor} Sensors are the interface between an agent and its environment, providing the raw (or interpreted) information the robot needs to understand its current context and make decisions. Perception of the environment encompasses multiple modalities, including vision through cameras, audition through microphones, touch through touch sensors, etc. Among these modalities, vision is critical. Most research uses vision as the only way for agents to perceive the environment.

\noindent \textbf{Actor.} Actors are the mechanisms through which an agent exerts influence or effectuates changes in its environment. They are the output devices that allow the agent to perform actions, such as motors for movement, robotic arms for manipulation, and communication interfaces for interaction with other systems or humans. The actions taken by the agent are determined by the decisions made within its planning system and are executed through the actuators.

\noindent \textbf{Planning.} Planning is the cognitive process that enables the autonomous agent to determine a sequence of actions that will lead to achieving its goals. It involves analyzing the current state of the environment as perceived by the sensors, defining the desired end state, and selecting the most appropriate actions to bridge the gap between the current and desired states. The planning component must consider the agent's capabilities, constraints, and the potential consequences of its actions. Effective planning allows the agent to act purposefully and adaptively, optimizing its behavior to achieve its objectives efficiently and effectively.

\noindent \textbf{World Model.} A world model is an internal representation of the surrounding environment. This model is crucial for the agent's ability to understand the context in which it operates, predict the outcomes of its actions, and make informed decisions. The world model interacts with the other three components through \textit{tell} and \textit{ask} interfaces\cite{sakagami2023robotic}. That is to say, it receives information from other components to update its state and also responds to queries from other components. A robust world model can reasonably predict the future state when told with current perceptions and actions, thereby guiding the planning component to make wiser decisions.

\subsubsection{Widely-used Model Structure}
A world model's key ability is predicting the environment's future state. Given the inherent randomness in most environments, predictions should maintain a balance between determinism and uncertainty. Many researches have been conducted on this problem, proposing a variety of model structures. Figure \ref{fig:wm_ag} shows the works in this field. Among these works, the most widely-used structures are RSSM\cite{hafner2019planet,deng2022dreamerpro,wu2023daydreamer,okada2022dreamingv2}, JEPA\cite{lecun2022jepa,assran2023ijepa,fei2024ajepa,bardes2023mcjepa,bardes2024vjepa}, and Transformer-based models\cite{chen2022transdreamer,zhang2023storm,robine2023twm,bruce2024genie, wang2024worlddreamer}.

\noindent \textbf{Recurrent State Space Model.} The Recurrent State Space Model (RSSM) is the core structure of the Dreamer series. RSSM aims to facilitate prediction in latent spaces. It learns a dynamic model of the environment from pixel observations and selects actions by planning in the encoded latent space. By decomposing the latent state into stochastic and deterministic parts, this model considers both deterministic and stochastic factors of the environment. Due to its exceptional performance in continuous control tasks for robots, many subsequent works have expanded upon its foundation.

\noindent \textbf{Joint-Embedding Predictive Architecture.} The Joint-Embedding Predictive Architecture (JEPA) is proposed in a paper by LeCun\cite{lecun2022jepa} that laid out a conceptual framework for future autonomous machine intelligence architecture. It learns the mapping from input data to predicted output. This model is different from traditional generative models as it does not directly generate pixel-level output, but makes predictions in a higher-level representation space, allowing the model to focus on learning more semantic features. Another core idea of JEPA is to train the network through self-supervised learning so that it can predict missing or hidden parts in the input data. Through self-supervised learning, models can be pre-trained on a large number of unlabeled data and then fine-tuned on downstream tasks, thereby improving their performance on a variety of visual\cite{assran2023ijepa,bardes2023mcjepa,bardes2024vjepa} and non-visual tasks\cite{fei2024ajepa}.

\noindent \textbf{Transformer-based World Models.} The Transformer\cite{vaswani2017attention} originates from the natural language processing task. It operates on the principle of the attention mechanism, which enables the model to simultaneously focus on different parts of the input data. Transformers have been proven to be more effective than Recurrent Neural Networks (RNNs) in many domains that require long-term dependencies and direct memory access for memory-based reasoning\cite{banino2020memo}, thus gaining increasing attention in the field of reinforcement learning in recent years. Since 2022, multiple works have attempted to construct world models based on the Transformer and its variants\cite{micheli2023iris,robine2023twm,zhang2023storm}, achieving better performance than the RSSM model on some complex memory interaction tasks\cite{chen2022transdreamer}. Among them, Google's Genie\cite{bruce2024genie} has attracted considerable attention. This work constructs a generative interactive environment based on the ST-Transformer\cite{xu2021sttransformer}, trained through self-supervised learning from a vast collection of unlabeled internet video data. Genie demonstrates a new paradigm for manipulable world models, offering a glimpse into the immense potential for the future development of world models.

\subsection{Agents for Different Tasks}\label{sec:ag_task}
Many researchers have explored the application of agents in various fields and tasks, such as gaming, robotics, navigation, task planning, etc. Among the most widely studied tasks are games and robotics.

\subsubsection{Game Agent}
Getting AI systems to learn to play games has been an interesting topic for a long time. The research on game agents not only improves the game experience but more importantly, helps people develop more advanced algorithms and models.

With the introduction of the Arcade Learning Environment (ALE)\cite{bellemare2013arcade}, Atari games have gained a lot of emphasis as a benchmark for reinforcement learning. The Atari collection includes more than 500 games, covering a wide variety of game types and challenges, making it ideal for evaluating the capabilities of reinforcement learning algorithms. Many studies have shown that reinforcement learning can make agents play games at a level comparable to that of human players\cite{schulman2017ppo,hessel2018rainbow,espeholt2018impala,laskin2020curl}. However, most of them require a huge amount of interaction steps with the environment. World models can predict future states of the environment, allowing agents to learn in imagination, thus significantly reducing the number of interactions required for learning.

RES\cite{chiappa2017recurrent} is an RNN-based environment simulator that can predict the subsequent state of the environment based on a series of actions and corresponding environmental observations. Based on this capability, SimPLe\cite{kaiser2020simple} designs a novel stochastic video prediction model, which achieves significant improvement in sample efficiency. Under the constraint of 100K interactions, SimPLe has a much better performance in Atari games compared to previous model-free reinforcement learning methods.

DreamerV2\cite{hafner2021dreamerV2} trains a game agent based on the RSSM model\cite{hafner2019planet}. Unlike previous approaches that use continuous latent representations, DreamerV2 uses discrete categorical variables. This discretization method enables the model to capture the dynamic changes in the environment more accurately. DreamerV2 further uses the actor-critic algorithm to learn the behaviors purely from imagined sequences generated by the world model and achieves performance comparable to human players on the Atari 200M benchmark\cite{mnih2015dqn}.

IRIS\cite{micheli2023iris} is one of the pioneers that apply Transformer\cite{vaswani2017attention} in the world model. The agent learns its skill in a world model based on the autoregressive Transformer. As pointed out by Robine et al.\cite{robine2023twm}, the autoregressive Transformer can model more complex dependencies by allowing the world model to directly access previous states, while previous works can only view a compressed recurrent state. IRIS shows that the Transformer architecture is more efficient in sampling, outperforming humans in the Atari100k benchmark\cite{kaiser2020simple} by only two hours of gameplay.

TWM\cite{robine2023twm} proposes a Transformer-XL\cite{dai2019transformerxl}-based world model. Transformer-XL solves the problem of capturing long-distance dependencies in language modeling tasks by introducing the segment-level recurrence mechanism to extend the length of context. TWM migrates this capability into the world model, enabling the capture of long-term dependencies between the states of the environment. To run more efficiently, TWM further trains a model-free agent in the latent imagination, avoiding a full inference of the world model in runtime.

STORM\cite{zhang2023storm} sets a new record in no-resorting-to-lookahead-search methods on Atari100k benchmark\cite{kaiser2020simple} by stochastic Transformer. Inspired by the fact that introducing random noise into the world model helps enhance the robustness and reduce cumulative errors in autoregressive predictions, STORM employs a categorical variational autoencoder\cite{kingma2013auto}, which inherently has a stochastic nature.

Genie\cite{bruce2024genie} is a novel generative environment developed by the DeepMind team. It learns the ability to generate interactive 2D worlds by unsupervised learning from many internet videos without labels. The most attractive point is that it can not only generate an entirely new virtual environment based on image or text prompts but also predict coherent video sequences of that environment frame by frame based on user input actions. Genie enhances the efficiency of virtual content creation as well as provides a rich interactive learning platform for the training of future AI agents. Although the current video quality and frame rate still need improvement, it has already demonstrated the immense potential of generative AI in building future virtual worlds.

\subsubsection{Robotics}
Getting an agent to learn to manipulate a robot is a long-term challenge. Agents are desired to plan autonomously, make decisions, and control actuators (e.g., robotic arms and legs) to complete complex interactions with the physical world. Common basic tasks include walking, running, jumping, grasping, carrying, and placing objects. Some of the more complicated tasks require a combination of several basic tasks, such as taking a specific item out of a drawer or making a cup of coffee. 

One difference between a robot and a game agent is that the goal of the robot is to interact with the real environment, which not only makes the environment dynamics more complex and stochastic but also greatly increases the cost of interacting with the environment during the training process. Therefore, it is particularly important to reduce the number of interaction steps with the environment and enhance sampling efficiency in such scenarios. In addition, the control of the actuators is in a continuous action space, which is also very different from the discrete action space in the game environment.

Previous works of model-based planning\cite{deisenroth2011pilco,gal2016improving,henaff2018model} learn low-dimensional environment dynamics by assuming that access to the underlying state and the reward function is available. But in complex environments, this assumption is often untenable. Hafner et al.\cite{hafner2019planet} suggested learning the environment dynamics from pixels and planning in latent spaces. They proposed RSSM, which is the base of the later Dreamer-like world models. They achieved similar performance to the state-of-art model-free methods within less than 1/100 episodes on six continuous control tasks of DeepMind Control Suite (DMC)\cite{tassa2018dmc}, which proves that learning latent dynamics of the environments in the image domain is a promising approach.

However, PlaNet\cite{hafner2019planet} learns the behaviors by online planning, i.e., considering only rewards within a fixed imagination horizon, which brings shortsighted behaviors. To solve this problem, Hafner et al. further proposed DreamerV1\cite{hafner2020dreamerV1}, an agent that learns long-horizon behaviors purely from the imagination of the RSSM-based world model. Predicting in latent space is memory efficient, thus allowing imagine thousands of trajectories in parallel. DreamerV1 uses a novel actor-critic algorithm to learn behaviors beyond the horizon. The evaluation performed on visual control tasks of DMC shows that DreamerV1 exceeds previous model-based and model-free approaches in data efficiency, computation time, and final performance.

SafeDreamer\cite{huang2024safedreamer} aims to address safe reinforcement learning, especially in complex scenarios such as vision-only tasks. SafeDreamer employs an online safety-reward planning algorithm for planning within world models to meet the constraints of vision-based tasks. It also combines Lagrangian methods with online and background planning within world models to balance long-term rewards and costs. SafeDreamer demonstrates nearly zero-cost performance across low-dimensional and visual input tasks and outperforms other reinforcement learning methods in the Safety-Gymnasium benchmark\cite{ji2023safety}, showcasing its effectiveness in balancing performance and safety in reinforcement learning tasks.

The above works only learn and evaluate their performance in simple simulation environments, while the real environments often contain task-unrelated visual distractions such as complex backgrounds and varying lights. RSSM learns the world model by reconstructing image observations, making it very sensitive to visual distractions in images and difficult to capture small but important content. Therefore, based on DreamerV1\cite{hafner2020dreamerV1}, Dreaming\cite{okada2021dreaming} avoids the auto-encoding process by directly imagining and planning in the latent space, and trains the world model by contrastive learning, which does not rely on pixel-level reconstruction loss, so that the method is robust to visual distractions in the environment. DreamingV2\cite{okada2022dreamingv2} further explores how to apply contrastive learning to the discrete latent space of DreamerV2\cite{hafner2021dreamerV2}. Experimental results on 5 simulated robot tasks with 3D space and photorealistic rendering show that DreamingV2 can effectively handle complex visual observations, and the performance is significantly better than that of DreamerV2.

Similar efforts are made by DreamerPro\cite{deng2022dreamerpro} and Dr.G\cite{ha2023dream}, both of which use a reconstruction-free approach to address the visual sensitivity issue of RSSM. The difference is that DreamerPro uses the prototype learning method to train the prediction of the world model in the latent space, which avoids the expensive computation caused by the large batch size required for contrast learning. Dr.G, on the other hand, uses a self-supervised method of double-contrast learning to replace the reconstruction loss in DreamerV1\cite{hafner2020dreamerV1}. Both are evaluated in environments from DMC synthesized with complex background videos, verifying their robustness to visual distractions.

Besides those works involving simulated environments only, some works are trying to train a robot in the real world. The most difficult thing is that interaction with the real world is expensive or even dangerous. Thus the ability of \textit{training in imagination} is especially important in such scenarios. RobotDreamPolicy\cite{piergiovanni2019learning} learns a world model first and then learns the policy in the world model to reduce the interactions with the real environment. During the training of the world model, the robot executes random actions in the environment, collecting pairs of the image before action, action, and the image after action as the training data. DayDreamer\cite{wu2023daydreamer} applies DreamerV2\cite{hafner2021dreamerV2} to 4 real robots and directly trains the model online in real environments. The authors found in experiments that the Dreamer model is capable of online learning in the real world, and can master a skill in a very short time. These works provide strong evidence that the sample efficiency of the world model can help robots learn various skills efficiently with fewer interactions.

\subsubsection{Diverse Environments and Tasks}
Besides game and robotic tasks, some research works have looked at other tasks such as navigation. PathDreamer\cite{koh2021pathdreamer} applies the idea of world model to indoor navigation tasks. The world model is used to enhance environmental awareness and predictive planning. Given one or more previous observations, PathDreamer can predict plausible panoramic images of the future, even for unseen rooms or regions behind corners. Furthermore, PathDreamer innovatively uses 3D point clouds for environment representation, which significantly improves navigation success. 

The JEPA series of work applies the architecture proposed by LeCun\cite{lecun2022jepa} to a variety of modal understanding and prediction tasks. I-JEPA\cite{assran2023ijepa} is a non-generative self-supervised learning method that learns highly semantic visual representations by predicting the representations of different target blocks within the same image from a single context block. A-JEPA\cite{fei2024ajepa} proposes a self-supervised learning method based on audio spectrograms, which effectively applies the successfully masked modeling principle from the visual domain to audio. A context encoder is used to predict and align the representations of different target blocks from the same audio spectrogram. MC-JEPA\cite{bardes2023mcjepa} is a self-supervised learning method that simultaneously learns video content features and motion features through JEPA, using a shared encoder to improve the accuracy of motion estimation and enrich the content features to include motion information. V-JEPA\cite{bardes2024vjepa} extends I-JEPA to feature prediction in videos. It presents a suite of vision models that are exclusively trained based on the objective of feature prediction. These models are developed without relying on supervisory signals such as pre-trained image encoders, negative examples, text, and reconstruction techniques.


Other research efforts aim to study agents suitable for diverse tasks. DreamerV3\cite{hafner2023dreamerv3} is a universal algorithm that realizes cross-domain learning with fixed hyperparameters by signal amplitude transformation and robust normalization. The authors evaluated multiple benchmark sets from Atari games, high/low dimensional continuous control tasks, survival tasks, spatial and temporal reasoning tasks, etc. The results show that DreamerV3 can master different domains only by relying on the same set of hyperparameters, and its performance is even better than some specialized algorithms designed for specific domains. DreamerV3 is also the first agent to successfully collect diamonds from scratch in Minecraft without providing any human experience.

Plan2Explore\cite{sekar2020planning} proposes a self-supervised two-stage learning process. In the first stage, the agent explores the environment in a self-supervised manner, gathers information about the environment, and summarizes past experiences in the form of a parametric world model. It is worth noting that no reward information is provided to the agent during this phase, and the exploration is performed by the agent autonomously. Then the agent learns behaviors in the trained world model for specific tasks. This stage can be done with little or no interaction with the environment. The two-stage learning process allows the agent to obtain a more universal world model, making the agent learn downstream tasks more efficiently.

SWIM\cite{mendonca2023structured} aims to solve the learning of complex and general skills in the real world. SWIM claims that an agent must utilize internet-scale human video data to understand rich interactions carried out by humans and gain meaningful affordances. To this end, SWIM proposes a high-level, structured, human-centric action space that is applicable for both humans and robots. The world model is first trained from a large dataset containing around 50K egocentric videos. Then the world model is finetuned with robot data to fit the robot domain. After that, behaviors for specified tasks can be learned in the trained world model using the standard cross-entropy method\cite{rubinstein1999cem}. With the help of human action videos, SWIM achieves about two times higher success than prior approaches while requiring less than 30 minutes of real-world interaction data.

HarmonyDream\cite{ma2024harmonydream} identifies the world model as a multi-task model consisting of observation modeling tasks and reward modeling tasks. HarmonyDream argues that traditional world modeling methods, which tend to focus on observation modeling, can become difficult and inefficient due to the complexity of the environment and the limited capacity of the model. HarmonyDream maintains a balance between observation modeling and reward modeling by automatically adjusting the loss coefficient, which can be adapted to different types of tasks and avoid complicated hyperparameter adjustments.

RoboDreamer\cite{zhou2024robodreamer} learns compositional world models to enhance robotic imagination. It decomposes the video generation process and leverages the inherent compositionality of natural language. In this way, it can synthesize video plans of unseen combinations of objects and actions. RoboDreamer dissects language instructions into a set of primitives, which then serve as distinct conditions for a set of models to generate videos. This method not only demonstrates strong zero-shot generalization capabilities but also shows promising results in multimodal-instructional video generation and deployment on robotic manipulation tasks.

UniSim\cite{yang2024unisim} is a generative simulator for real-world interactions. UniSim contains a unified generative framework taking action as input that integrates diverse datasets across different modulations. With this approach, UniSim can simulate the visual outcomes of both high-level instructions and low-level controls. UniSim can be utilized for various applications, such as controllable game content creation and the training of embodied agents in simulated environments, which can be directly deployed in the real world.

\subsection{Commonly Used Benchmarks}\label{sec:ag_benchmark}

\begin{table}[tbp]
  \centering
  \caption{Human-normalized scores (HNS in \%) of world-model-based agents on 26 games in the Atari100k benchmark\cite{kaiser2020simple}. The highest score of each row is bolded.}
  \resizebox{\columnwidth}{!}
  {
    \begin{tabular}{l|c|c|c|c|c|c|c}
    \toprule
      & \thead{SimPLe \\ \cite{kaiser2020simple}} & \thead{DreamerV3 \\ \cite{hafner2023dreamerv3}} & \thead{IRIS \\ \cite{micheli2023iris}}
      & \thead{TWM \\ \cite{robine2023twm}}       & \thead{STORM \\ \cite{zhang2023storm}}          & \thead{Harmony \\ Dreamer \\ \cite{ma2024harmonydream}}  & \thead{DIAMOND \\ \cite{globerson2024diamond}}\\
    \midrule
      Alien          & 5.6   & 10.6  & 2.8            & 6.5            & \textbf{11.0}  & 9.6            & 7.5\\
      Amidar         & 4.0   & 7.8   & 8.0            & 6.8            & 11.6           & 7.9            & \textbf{12.8}\\
      Assault        & 58.7  & 93.1  & 250.6          & 88.6           & 111.4          & 150.2          & \textbf{251.0}\\
      Asterix        & 11.1  & 8.7   & 7.8            & 10.9           & 9.9            & 11.2           & \textbf{42.1}\\
      BankHeist      & 2.7   & 85.9  & 5.3            & 61.2           & 84.8           & \textbf{142.8} & 0.7\\
      BattleZone     & 4.8   & 28.4  & 30.8           & 7.8            & 32.1           & \textbf{40.5}  & 6.7\\
      Boxing         & 64.2  & 649.2 & 583.3          & 645.0          & 665.8          & 665.8          & \textbf{723.3}\\
      Breakout       & 51.0  & 101.7 & 284.7          & 63.5           & 49.7           & 178.1          & \textbf{454.2}\\
      Chopper.       & 2.6   & -5.9  & 11.5           & 13.5           & \textbf{16.4}  & 10.6           & 8.5\\
      CrazyClimber   & 206.8 & 345.0 & 193.8          & 243.7          & 223.5          & 287.3          & \textbf{352.9}\\
      DemonAttack    & 3.1   & 8.3   & \textbf{103.5} & 10.9           & 0.7            & 2.8            & 7.5\\
      Freeway        & 56.4  & 0.0   & 105.1          & 82.1           & \textbf{114.9} & 0.0            & 112.5\\
      Frostbite      & 4.0   & 19.8  & 4.5            & \textbf{33.0}  & 29.3           & 14.4           & 4.9\\
      Gopher         & 15.7  & 161.1 & 91.8           & 65.8           & 370.4          & \textbf{593.3} & 261.7\\
      Hero           & 5.5   & 34.0  & 20.2           & 20.9           & 33.6           & \textbf{41.4}  & 15.4\\
      Jamesbond      & 26.1  & 151.9 & 158.4          & 121.8          & \textbf{175.3} & 105.2          & 145.5\\
      Kangaroo       & 0.0   & 135.6 & 26.4           & 39.8           & 139.3          & 169.8          & \textbf{178.7}\\
      Krull          & 56.8  & 579.3 & 470.1          & 445.1          & 638.4          & 576.7          & \textbf{656.9}\\
      KungFuMaster   & 65.0  & 94.1  & 95.7           & 108.1          & \textbf{115.3} & 97.9           & 82.1\\
      MsPacman       & 17.6  & 15.3  & 10.4           & 19.3           & \textbf{35.6}  & 20.7           & 24.8\\
      Pong           & 94.9  & 109.6 & 100.0          & 111.9          & 89.8           & 112.5          & \textbf{116.4}\\
      PrivateEye     & 0.0   & 1.2   & 0.1            & 0.1            & \textbf{11.2}  & 4.2            & 0.1\\
      Qbert          & 8.5   & 24.4  & 4.4            & 23.8           & \textbf{32.8}  & 28.4           & 32.6\\
      RoadRunner     & 71.9  & 198.6 & 122.6          & 116.1          & 224.1          & 186.8          & \textbf{263.8}\\
      Seaquest       & 1.5   & 1.3   & 1.4            & \textbf{1.7}   & 1.1            & 1.4            & 1.1\\
      UpNDown        & 25.2  & 63.9  & 27.0           & \textbf{138.4} & 66.8           & 92.7           & 29.8\\
    \midrule
      Avg. HNS & 33.2 & 112.4 & 104.6 & 95.6 & 126.7 & 136.6&\textbf{145.9}\\
    \bottomrule
    \end{tabular}%
  }
  \label{tab:atari100k}%
\end{table}%

\begin{table}[tbp]
  \centering
  \caption{The performance (Episode Return) of some world-model-based agents in different robot tasks. This table is reported by Okada et al.\cite{okada2022dreamingv2}. Episode Return is defined as the sum of all rewards earned by the agent in a full episode. The highest value of each row is bolded.}
  \resizebox{0.99\columnwidth}{!}
  {
    \begin{tabular}{l|c|c|c|c|c}
    \toprule
          &  \thead{Dreamer \\ \cite{hafner2020dreamerV1}} & \thead{DreamerV2 \\ \cite{hafner2021dreamerV2}} & \thead{Dreaming \\ \cite{okada2021dreaming}} &\thead{DreamingV2 \\ \cite{okada2022dreamingv2}} & \thead{DreamerPro \\ \cite{deng2022dreamerpro}} \\
    \midrule
     \multicolumn{6}{l}{3D Robot-arm tasks from Dreaming\cite{okada2021dreaming} and RoboSuite\cite{zhu2022robosuite}}\\
    \midrule
    UR5-reach &  701±223 & 704±222 & 752±1178 &\textbf{776}±194 & 668±252 \\
    Lift  &  134±46 & 165±126 & 174±107 &\textbf{327}±150 & 138±64 \\
    Door  &  154±32 & 190±126 & 319±173 &\textbf{383}±143 & 111±110 \\
    PegInHole &  354±47 & 376±59 & 353±50 &\textbf{436}±26 & 327±43 \\
    \midrule
     \multicolumn{6}{l}{Difficult pole-swingup tasks from DMC\cite{tassa2018dmc}}\\
    \midrule
    Acrobot-swingup &  382±147 & 309±131 & 359±111 &\textbf{470}±129 & - \\
    Cartpole-two-poles &  256±65 & 248±103 & 273±53 &\textbf{308}±55 & - \\
    \midrule
     \multicolumn{6}{l}{3D robot tasks from DMC\cite{tassa2018dmc}}\\
    \midrule
    Quadruped-walk &  242±120 & 350±89 & 379±189 &\textbf{492}±127 & - \\
    Quadruped-run &  269±114 & 352±68 & 339±128 &\textbf{385}±91 & - \\
    Reach-duplo &  5±11  & 149±62 & 145±61 &\textbf{199}±43 & 87±76 \\
    \midrule
     \multicolumn{6}{l}{2D robot tasks from DMC\cite{tassa2018dmc}}\\
    \midrule
    Cheetah-run &  776±120 & \textbf{811}±75 & 542±132 &768±24 & - \\
    Walker-walk &  906±70 & \textbf{951}±28 & 518±76 &857±115 & - \\
    Reacher-easy &  658±429 & 923±215 & \textbf{947}±100 &924±210 & - \\
    Reacher-hard &  247±392 & 175±340 & \textbf{743}±346 &598±447 & - \\
    Finger-turn-easy &  665±430 & 498±469 & \textbf{842}±286 &434±469 & - \\
    Finger-turn-hard &  533±426 & 600±417 & \textbf{858}±210 &484±434 & - \\
    \bottomrule
    \end{tabular}%
  }
  \label{tab:dmc}%
\end{table}%

A variety of benchmarks are used to measure the performance of game agents and robotics. The evaluation method is usually to test the completion of several specific tasks or the rewards obtained by the agent after a limited amount of interactive learning in a specific environment.

Atari100k\cite{kaiser2020simple} is the most commonly used benchmark for game agents, which uses a subset of 26 Atari games from the Arcade Learning Environment\cite{bellemare2013arcade}. For each game, the agent is allowed to collect up to 100K interactions. With 4 frames per interaction, this is equivalent to 400K frames or 114 minutes (at 60FPS). To normalize the scores across different games, a metric called Normalized Human Score (NHS) \cite{mnih2015dqn} is proposed, which is defined as:
\begin{equation}
    NHS=\frac{score_{\text{agent}} - score_{\text{random}}}{score_{\text{human}} - score_{\text{random}}}
\end{equation}
Where $score_{\text{human}}$ is the score achieved by the professional human player and $score_{\text{random}}$ is the score achieved by an agent using purely random policy. This metric evaluates the performance of agents compared to the professional human player. Table \ref{tab:atari100k} collects the performance reported by the world model-based game agents mentioned in this survey. Overall, recent methods have been able to outperform human players in about half of these 26 games with a constraint of only 100K interactions, and in some games, several times over. At the same time, in other games such as Alien, Amidar, and Seaquest, they perform much worse than human players. This may be because the environment dynamics of these games are more complex, and 100K interactions are not enough for the agent to have a full understanding of the environment. On the other hand, low-quality images make some important elements easily ignored by image-reconstruction-based algorithms, resulting in an incorrect understanding of the environment.

For robotic tasks, there are several benchmarks adopted for different tasks and environments. DMC\cite{tassa2018dmc} is the most commonly used benchmark for robot learning. It contains a virtual environment that supports research into how agents learn complex physical tasks. This environment offers a diverse set of control tasks, from simple object moving to complex manipulator operations, as well as navigation tasks in 3D space. These tasks are built on top of the MuJoCo physics engine\cite{mujoco}. It also supports high-dimensional observing spaces, including pixel-level visual inputs, which makes it suitable for studying vision-driven reinforcement learning algorithms. To increase the visual diversity, DMC Remaster\cite{grigsby2020dmcr} extends DMC with seven types of visual factors, including the ground texture, the background, the color of robot, the color of target, the specular property, the camera position, and the light, thus presents a greater challenge to the visual robustness of the algorithm.

Another common benchmark is RoboSuite\cite{zhu2022robosuite}. It is a robot learning simulation framework powered by MuJoCo physics engine that provides a standardized benchmark environment for robot learning research. The RoboSuite includes a variety of robot models, grippers, controller modes, and a standardized set of benchmark tasks. In addition, it supports generating new environments programmatically with a modular API design that allows researchers the flexibility to design new robotic simulation environments.

Other benchmarks for robotic tasks include Meta-World\cite{yu2020metaworld} which contains 50 distinct robotic manipulation tasks for meta-RL and multi-task learning, RLBench\cite{james2020rlbench} which contains 100 unique, hand-designed tasks, covering everything from simple goal-reaching and door opening to more complex multi-stage tasks.

Due to the choice of different tasks and interaction constraints in different works, the results of robotic research works are difficult to align. DreamingV2\cite{okada2022dreamingv2} evaluates a relatively complete set of these works, which covers discrete/continuous latent space and with/without image reconstruction. We refer to their evaluation results in this survey, which are presented in Table \ref{tab:dmc}. The experiment analyzes the impact of two factors, the discreteness or continuity of the latent space and the presence or absence of image reconstruction, on the learning effectiveness of the agent.

\section{Discussion}
Despite the recent surge in research on general world models~\cite{ha2018worldmodel,bruce2024genie} and specific applications in areas like autonomous driving~\cite{gaia, drivedreamer, drivewm, wovogen, occworld, copilot4d, driveworld, adriver-i, genad} and robotics~\cite{hafner2020dreamerV1, hafner2021dreamerV2, hafner2023dreamerv3, wu2023daydreamer}, numerous challenges and opportunities await further exploration.
In this section, we delve into the intricate challenges faced by general world models and their current technical constraint, alongside envisioning potential future directions for their development.
Additionally, we explore the unique challenges and promising avenues in the fields of autonomous driving and autonomous agents.
Furthermore, we reflect on the ethical and safety considerations arising from the deployment of these models.

\subsection{General World Models}

General world models aim to represent and simulate a wide range of situations and interactions, like those encountered in the real world. Recent advancements in generative models have greatly improved video generation quality. Notably, Sora can create high-definition videos up to one minute in length, closely mimicking the physical world, showing great potential for general world models. However, it's crucial to address existing issues and challenges for future progress.

\subsubsection{Challenges}
Video generation is not synonymous with world models. While video generation may serve as one manifestation of world models, it does not fully address the core challenges inherent to world models. We will discuss several challenges that we deem important for world models in the following.

\begin{figure}[t]
	\centering
	\includegraphics[width=0.95\linewidth]{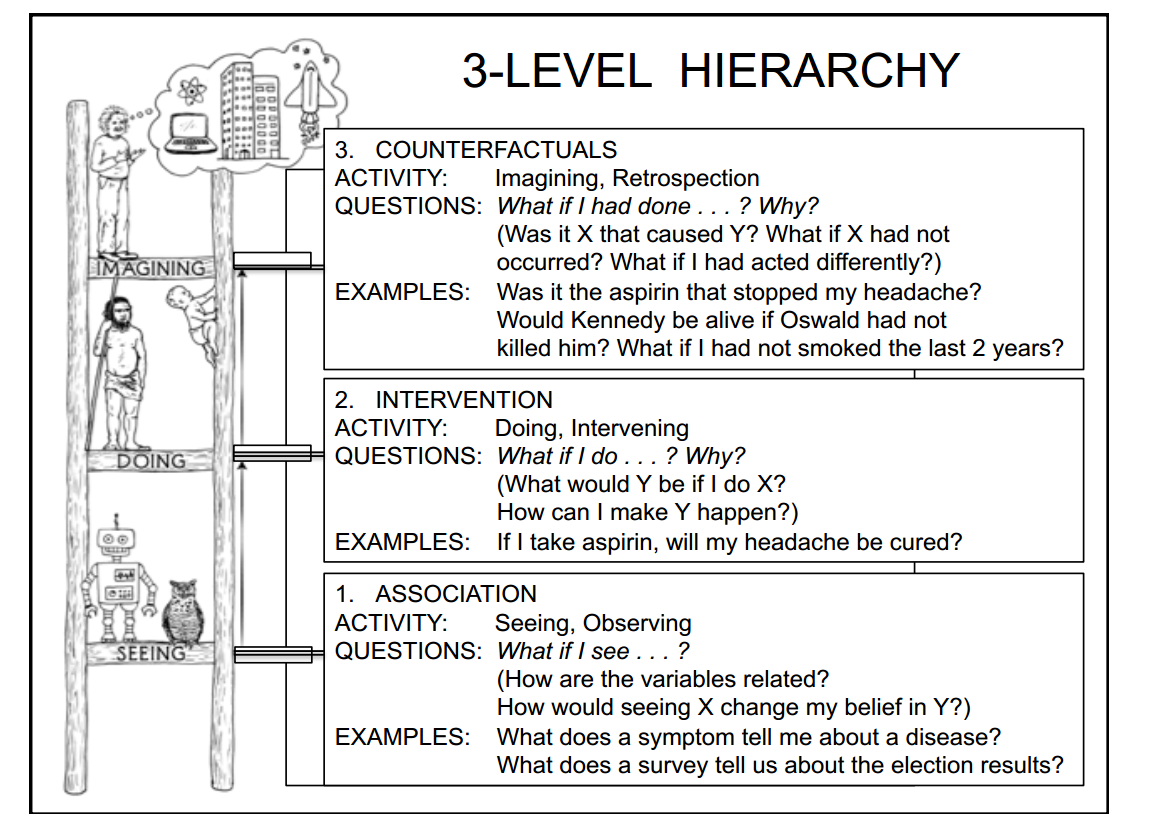} 
	\caption{The three level hierarchy of intelligence~\cite{pearl2018book}. World models are expected to conduct counterfactual reasoning.}
	\label{fig:reasoning}
\end{figure}

\noindent \textbf{Causal Reasoning.}
As a predictive model, the essence of world modeling lies in its capacity for reasoning. The model should be capable of inferring outcomes of decisions never encountered before, rather than solely making predictions within known data distributions.
As discussed in ~\cite{pearl2018book} and illustrated in Figure~\ref{fig:reasoning}, we expect world models to possess the ability of counterfactual reasoning, whereby outcomes are inferred through rational imagining. This ability is inherently human but remains a challenging task for current AI systems. For example, imagine an autonomous vehicle facing a sudden traffic accident or a robot in a new environment. A world model with counterfactual reasoning can simulate different actions they could take, predict outcomes, and choose the safest response—even in new situations. This would significantly improve autonomous systems' decision-making, helping them handle new and complex scenarios.


\noindent \textbf{Physical Laws.}
Although Sora's video generation is impressive, it's argued to fall short as a world model because it doesn't fully comply with physical laws. Realism, seen in Sora's videos, isn't the same as reality, which demands strict obedience to physical laws like gravity, light interaction, and fluid dynamics. While Sora has improved in modeling movement, including pedestrians and rigid body movements~\cite{esser2023structure}, it still struggles with accurately simulating fluids and complex physical phenomena. Training Sora with just video-text pairs isn't enough to grasp these complexities. Understanding physical laws often requires specific observations, suggesting that combining Sora with physics-driven simulators could be beneficial. Although these simulators might not achieve Sora's level of realism, they correctly follow physical properties.

\noindent \textbf{Generalization.}
Generalization capability is a crucial aspect of world models, emphasizing not only data interpolation but, more importantly, data extrapolation. For instance, in autonomous driving, real-life accidents or abnormal driving behaviors are rare occurrences. Therefore, can the learned world model imagine these rare driving events? This requires the model to move beyond simply memorizing the training data and instead develop a robust understanding of the underlying principles governing driving dynamics and road scenarios. By extrapolating from known data and simulating a wide range of potential situations, the world model can better prepare autonomous vehicles to navigate safely in the real world, even in unfamiliar or unexpected circumstances.

\noindent \textbf{Computational Efficiency.}
Efficiency in video generation is currently a significant limitation. To maintain consistency in video generation, autoregressive methods are often employed, leading to a considerable increase in generation time. 
Based on the news and anaylysis in the internet, Sora may take over about one hour to generate a video with one minute length.
Although a series of distillation-based methods~\cite{luo2023latent, chen2024pixart} have emerged in image generation, yielding significant acceleration in performance, research in the field of video generation remains limited.

\noindent \textbf{Evaluation System.}
Current world models are predominantly based on generative model research, with evaluation metrics primarily focusing on the quality of generation, such as FID~\cite{heusel2017gans} and FVD~\cite{unterthiner2018towards}. Additionally, there are some works proposing more comprehensive evaluation benchmarks, such as CLIPScore~\cite{hessel2021clipscore}, T2VScore~\cite{wu2024towards}, VBench~\cite{huang2023vbench}, EvalCrafter~\cite{liu2023evalcrafter}, PEEKABOO~\cite{jain2023peekaboo}, and others.
However, generation metrics alone cannot reflect the predictive rationality of world models.
This highlights the need for human-centric evaluation~\cite{cho2024sora}, which measures whether the generated videos meet users' expectations or align with human reasoning.
By incorporating human feedback, evaluations become more comprehensive, considering realism, coherence, and relevance. This approach also offers insights into real-world utility, guiding further development and refinement for practical applications.

\subsubsection{Future Perspectives}
Despite the acclaimed success of recent world model studies, and considering some of the core challenges we discussed before, we believe that future research on world models can step further in the following directions.

\noindent \textbf{3D World Simulator.}
Video generation has advanced significantly in simulating various aspects of the world, but the world exists fundamentally in three dimensions. Therefore, future world models should possess the capability to predict and comprehend 3D spatial environments. This involves not only capturing the visual appearance of objects and scenes but also encoding their spatial relationships, depth information, and volumetric properties. Extending world models into three-dimensional space can enable more immersive and realistic simulations, facilitating applications in virtual reality~\cite{lee2021all}, augmented reality, robotics, and autonomous systems. Moreover, 3D world models can enhance the ability to interpret and interact with the physical world.

\noindent \textbf{World Models for Embodied Intelligence.}
World models for embodied intelligence~\cite{vemprala2024chatgpt} involve creating comprehensive representations of the environment that an agent interacts with. This implies that world models can serve as simulators to train embodied agents' decision-making processes, as demonstrated by Drive-WM's~\cite{drivewm} preliminary attempts in the field of autonomous driving. Moreover, integration with embodied intelligence enriches their direct interaction with the environment, significantly enhancing machines' understanding of and adaptability to the physical world.

\subsection{World Models for Autonomous Driving}
While extensive research has been conducted on world models in autonomous driving, the current state of world models remains rudimentary compared to the comprehensive mental world models possessed by a skilled human driver. 
Significant challenges persist in areas like action controllability, 3D consistency, and overcoming data limitations. Nevertheless, we hold firm in the belief that the foundational model for autonomous driving will be based on world models, enabling effective interaction and comprehensive understanding of the physical world.

\subsubsection{Challenges}

\noindent 
\textbf{Action Controllability.}
In the realm of autonomous driving, the emphasis is on action-conditioned generation rather than text-conditioned video generation. While this area has garnered attention, only a handful of studies have delved into it. For instance, GAIA-1~\cite{gaia} and DriveDreamer~\cite{drivedreamer} focus on steering and throttle conditioning, while Drive-WM~\cite{drivewm} utilizes planning trajectories for better integration with end-to-end driving systems.
However, achieving fine-grained control over actions remains highly challenging. For instance, when attempting to control a vehicle to perform unconventional maneuvers such as high-speed turns or U-turns, the quality of generation noticeably deteriorates. This limitation is also influenced by the distribution of normal data. Actions, being continuous variables, pose difficulty in learning their latent space representations from limited data samples. Current methods are only capable of achieving coarse motion control, emphasizing the considerable gap that still exists towards achieving fine-grained control.

\noindent 
\textbf{3D Consistency.}
3D consistency is crucial for autonomous driving. Although current video generation techniques may appear realistic, ensuring their 3D consistency is challenging, thus compromising the reliability of world model generation. 
However, if the world model is to be truly applied, the ability to consistently generate 3D spaces must be further improved.
While the Sora team believes that scaling up can enable models to learn 3D consistency from videos, this implicit learning approach is obviously less secure for autonomous driving. Given the abundance of sensors in autonomous vehicles, world models can extend beyond mere video generation. For instance, conditioning on point clouds or occupancy grids can significantly enhance 3D consistency.

\noindent 
\textbf{Data Limitations.}
Data plays a crucial role in training foundation models. Unlike the readily available image and text data on the internet, autonomous driving encounters significant challenges in data collection, making world model construction exceedingly difficult.
Firstly, autonomous driving data collection differs substantially from human learning due to fixed sensor positions. Humans learn about the world's physics through passive observation and active interaction, while autonomous vehicles lack this flexibility. Understanding the consequences of the ego-agent's actions on the environment is vital for reasoning about interactions. However, such data is often scarce or hard to obtain, presenting a significant challenge in world model construction.
Secondly, privacy concerns and commercial competition often deter automotive companies from sharing their autonomous driving data. This not only limits the scale of available data but also restricts its diversity.
Lastly, data collection typically exhibits a long-tail distribution, emphasizing the importance of rare scenarios that are nonetheless crucial for autonomous driving. Therefore, the efficient selection of such data remains a challenging and unresolved issue. 
While GenAD~\cite{genad} has explored training world models using internet data, the effectiveness remains preliminary. Addressing these data limitation issues will facilitate research on autonomous driving world models.

\subsubsection{Future Perspectives}

\noindent 
\textbf{End-to-end Foundation Driving Models.}
The world model is crucial for building the end-to-end foundation model for autonomous driving. As a simulator of the real world, it can not only provide high-quality data but also enable a closed-loop training environment for decision-making.
Although the driving domain is more restricted compared to general scenarios, it involves rich interactions and an understanding of spatial and temporal information, which are currently lacking in text-based video generation models.
Currently, the world model for autonomous driving is still far from achieving this goal. The best model, GAIA-1~\cite{gaia}, is trained on 4,700 hours video data, akin to GPT predicting the next token. However, with a model size of 9B, it still falls far short compared to large language models.
However, the shift towards big data-driven autonomous driving is undoubtedly an inevitable trend. Models will increasingly comprehend reality and grasp the rules and techniques of driving from data, rather than relying solely on manually designed rules. In this regard, Tesla's FSD beta 12.3 has demonstrated amazing driving capabilities, offering a glimpse of hope for the future end-to-end foundation model of driving.

\noindent 
\textbf{Real-world Driving Simulators.}
While many end-to-end autonomous driving methods are under research in the CARLA simulator, the inherent disparities between simulated and real-world environments present significant challenges. This highlights the necessity of constructing more realistic real-world driving simulations in the future.
Leveraging the robust predictive capabilities of world models, we can create even more realistic driving simulators that extend beyond mere video generation. Such simulators must also focus on aspects like scene layout control, lighting control, and object manipulation. Furthermore, world models can be seamlessly integrated with previous simulation~\cite{yang2023unisim} efforts based on MVS~\cite{aa-rmvsnet,bmvs}, NeRF~\cite{nerf}, and 3D Gaussian Splatting~\cite{3dgs}, thereby enhancing the scene generalization capabilities of existing methods.
By utilizing more realistic driving simulators for model training, it can greatly facilitate the deployment of autonomous driving systems that perform reliably in practical settings~\cite{zhu2021deep}.

\subsection{World Models for Autonomous Agents}
Autonomous agents encompass both physical robots in the real world and intelligent agents in digital environments. World models have the capability to simulate not only the intricate complexities of the physical world but also the nuances of digital environments.
From the perspective of autonomous agents, world models present some new challenges and opportunities.

\subsubsection{Challenges}

\noindent \textbf{Understand the Environment Dynamics.}
Agents need to understand their environments to function effectively. For physical robots, this means grappling with the complex and often uncertain dynamics of the physical world, a task made difficult by limited observations and the probabilistic nature of real-world changes. Unlike robots, humans navigate this complexity well thanks to multisensory perception, genetic knowledge, and the ability to learn from experience and share knowledge.
To enhance an agent's understanding of its environment, we can draw inspiration from human capabilities in three ways: First, by enhancing multimodal perception, allowing agents to gather more comprehensive information through integrated models that encompass vision, sound, and touch. Examples of this approach include the development of large language models like GPT-4V~\cite{achiam2023gpt} and Gemini~\cite{team2023gemini}. Second, leveraging extensive internet data for unsupervised learning can aid agents in acquiring fundamental cognitive abilities. Lastly, advancing and disseminating sophisticated knowledge through systems like LeCun's multi-level knowledge induction \cite{lecun2022jepa} facilitates agents in rapidly attaining a deeper understanding of their environment.

\noindent \textbf{Task Generalization.}
Agents in real-world applications frequently encounter a range of diverse tasks, necessitating world models that can not only handle familiar tasks but also generalize effectively to novel, unseen ones. This task generalization capability is crucial for agents, yet current robots still face significant challenges in this regard. The majority of robots today are specialized models, tailored to perform specific functions such as sweeping, transporting, cooking, and the like, limiting their adaptability and versatility in handling a wider range of tasks.
This implies that learning world models cannot solely rely on imitation and generation; rather, it is essential to abstract common sense from diverse tasks. Such common sense enables agents to migrate and comprehend different tasks more easily. Mere reliance on big data learning is an inefficient and poorly generalizable approach. This is analogous to the concept of meta-learning, where meta-learning methods train agents to learn how to learn, enabling them to quickly adapt to new tasks. Additionally, multi-task learning frameworks empower agents to train on multiple tasks simultaneously, identifying and leveraging the commonalities between them.

\subsubsection{Future Perspectives}

\noindent \textbf{Knowledge Injection through Large Language Model.} 
LLM has demonstrated astonishing comprehension abilities over the past two years. Through the language learning, the model has acquired a certain amount of knowledge about the world.
Leveraging this accumulated knowledge, the LLM can serve as a prior for world models, enabling the model to learn different tasks more efficiently.
Just like humans, world models initially envision scenarios based on their preexisting knowledge and subsequently refine their understanding through feedback obtained from the actual environment.
We believe that the integration of world models with large language models represents one of the promising directions for future development.

\noindent \textbf{Real-world Application.}
While the Dreamer series of algorithms~\cite{hafner2020dreamerV1,hafner2021dreamerV2,hafner2023dreamerv3} has shown promise in learning from limited interaction through planning within simulated environments and gaming scenarios, their application in real-world robotics remains largely unexplored~\cite{wu2023daydreamer}. However, the transition from simulation to reality is an inevitable direction for future research. The real world introduces additional uncertainties, including observation errors and control precision, making it crucial to investigate the effectiveness of world models for physical robots in real-world settings.

\subsection{Ethical and Safety Concerns}
The main concerns surrounding tools like Sora revolve around their safety and ethical impacts. 

\noindent \textbf{Model Accountability.}
As a powerful predictive model, ensuring the reliability of world model predictions is a critical concern. Accountability measures are indispensable to validate the accuracy and fairness of model outputs, particularly considering their potential impact on decision-making processes.
For instance, in autonomous driving, the reliability of world model predictions is essential for ensuring safety. 
Moreover, accountability measures should also address issues of fairness, ensuring that world models do not exhibit biases~\cite{zhou2024bias} that could disproportionately impact certain groups or communities. 

\noindent \textbf{Disinformation.}
Hyper-realistic videos generated by visual generative AI present an alarming threat, particularly in their potential to create emotionally manipulative content that spreads misinformation, especially during critical events like elections. The proliferation of fabricated videos depicting politicians in fictitious scenarios significantly distorts public opinion. Furthermore, this misinformation seeps into education, making it harder to distinguish reality from falsehood. Tackling this issue demands collaborative action from governments, technology firms, media organizations, and civil society to establish a reliable information dissemination system.

\noindent \textbf{Data Privacy.}
The abundance of data undoubtedly propels the rapid development of large foundation models, but it also raises concerns about privacy protection. 
In a recent study~\cite{staab2023beyond}, mainstream large language models like GPT-4, Llama-2, and Claude-2 were used to infer specific privacy datasets.
It was found that these large models could automatically deduce various real privacy data hidden within the text content analyzed from users' texts alone.
Compared to textual data, the internet holds a vast amount of video data and continues to update daily, which requires even more attention to privacy concerns. Especially for large-scale models used in video generation, it is essential to disclose the sources of training videos to prevent personal privacy data from being unknowingly used for training. Additionally, corresponding laws and policies should be established to clarify the protection and requirements of personal privacy data.


\section{conclusion}
In this survey, we conduct a comprehensive review of general world models, underlining their pivotal importance in the pursuit of AGI and their fundamental applications across a myriad of domains, from immersive virtual environments to sophisticated decision-making systems. Through our examination, the emergence of the Sora model is highlighted for its unparalleled simulation capabilities and nascent understanding of physical principles, marking a significant milestone in the evolution of world models. We delve deeply into the current innovations, with a particular focus on the application of world models for video generation, autonomous driving, and the operation of autonomous agents.
Despite the progress and promising prospects, we also critically evaluate the challenges and limitations facing current world model methodologies, contemplating their complexity, ethical considerations, and scalability. This comprehensive review not only showcases the current state and potential of world models but also illuminates the path toward their future development and application. We hope this survey can inspire the community toward novel solutions, thereby broadening the horizon for world models and their applications in shaping the future of AGI.

{\scriptsize
\bibliographystyle{ieee}
\bibliography{egbib}
}

\end{document}